%% file: main.tex
\documentclass[runningheads]{llncs}

\usepackage{eccv}

\usepackage{eccvabbrv}

\usepackage{graphicx}
\usepackage{booktabs}
\usepackage{amsmath}
\usepackage{amssymb}

\usepackage{times}
\usepackage{epsfig}
\usepackage{comment}
\usepackage{epigraph}
\usepackage{multirow}
\usepackage{hhline}
\usepackage{makecell}
\usepackage{bm}
\usepackage{subcaption}
\usepackage{soul}
\usepackage{array}
\usepackage{color,colortbl}

\makeatletter
\def\blfootnote{\gdef\@thefnmark{}\@footnotetext}
\makeatother

\definecolor{beaublue}{rgb}{0.74, 0.83, 0.9}

\usepackage[accsupp]{axessibility}  % Improves PDF readability for those with disabilities.

\usepackage{hyperref}

\usepackage{orcidlink}

\begin{document}

\title{Learning by Aligning 2D Skeleton Sequences and Multi-Modality Fusion} 

\titlerunning{Learning by Aligning 2D Skeletons}

\author{Quoc-Huy Tran$^*$\orcidlink{0000-0003-1396-6544} \and
Muhammad Ahmed$^*$\orcidlink{0009-0009-6064-8929} \and
Murad Popattia\orcidlink{0009-0008-5286-3082} \and
M. Hassan Ahmed\orcidlink{0009-0001-2925-1984} \and
Andrey Konin\orcidlink{0000-0003-1886-2402} \and
M. Zeeshan Zia\orcidlink{0000-0001-8221-2637}}

\authorrunning{Tran et al.}

\institute{Retrocausal, Inc., Redmond, WA\\
\url{www.retrocausal.ai}}

\maketitle

\begin{abstract}
This paper presents a self-supervised temporal video alignment framework which is useful for several fine-grained human activity understanding applications. In contrast with the state-of-the-art method of CASA, where sequences of 3D skeleton coordinates are taken directly as input, our key idea is to use sequences of 2D skeleton heatmaps as input. Unlike CASA which performs self-attention in the temporal domain only, we feed 2D skeleton heatmaps to a video transformer which performs self-attention both in the spatial and temporal domains for extracting effective spatiotemporal and contextual features. In addition, we introduce simple heatmap augmentation techniques based on 2D skeletons for self-supervised learning. Despite the lack of 3D information, our approach achieves not only higher accuracy but also better robustness against missing and noisy keypoints than CASA. Furthermore, extensive evaluations on three public datasets, i.e., Penn Action, IKEA ASM, and H2O, demonstrate that our approach outperforms previous methods in different fine-grained human activity understanding tasks. Finally, fusing 2D skeleton heatmaps with RGB videos yields the state-of-the-art on all metrics and datasets. To our best knowledge, our work is the first to utilize 2D skeleton heatmap inputs and the first to explore multi-modality fusion for temporal video alignment. Our code and dataset are available on our research website: \url{https://retrocausal.ai/research/}.
\end{abstract}

\input{Sections/introduction.tex}
\input{Sections/relatedwork.tex}
\input{Sections/method.tex}
\input{Sections/experiments.tex}
\input{Sections/conclusion.tex}

\appendix
\input{supp.tex}

\bibliographystyle{splncs04}
\bibliography{references}
\end{document}

%% file: Sections/introduction.tex
\section{Introduction}
\label{sec:introduction}
{\blfootnote{$^*$ indicates joint first author. \{huy,ahmed,murad,hassan,andrey,zeeshan\}@retrocausal.ai.}} 

\begin{figure}[t]
	\centering
		\includegraphics[width=1.0\linewidth, trim = 0mm 70mm 0mm 0mm, clip]{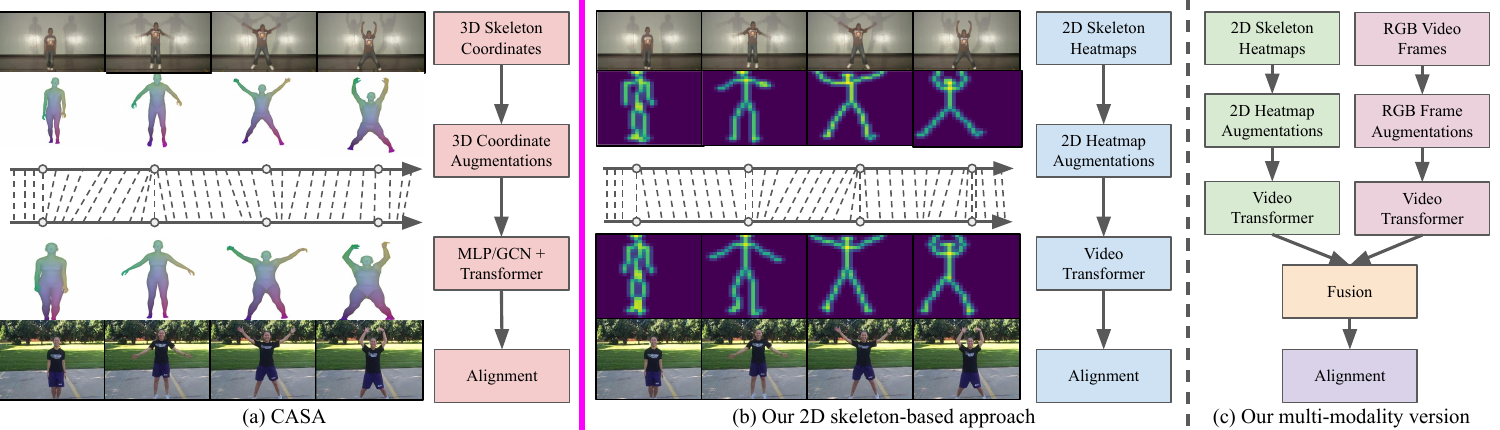}
	\caption{(a) The state-of-the-art method in self-supervised learning, i.e., CASA, uses 3D skeleton-based temporal video alignment as a pretext task and performs 3D skeleton augmentations. (b) Our approach relies on 2D skeleton-based temporal video alignment and conducts 2D skeleton augmentations. We use 2D skeleton heatmaps, which are fed to a video transformer for learning useful spatiotemporal and contextual features. Our method obtains higher accuracy and better robustness against missing and noisy keypoints, while showing superior performance in various fine-grained human activity understanding tasks. (c) We further fuse 2D skeleton heatmaps with RGB videos, establishing the state-of-the-art across all metrics and datasets.}
	\label{fig:teaser}
\end{figure}

% Temporal video alignment, challenges, and applications
We study the problem of temporal video alignment~\cite{sermanet2018tcn,dwibedi2019tcc,haresh2021learning,liu2022learning,kwon2022context}, which aims to find dense framewise correspondences between two videos capturing the same activity, despite discrepancies in human appearance, background clutter, human motion, and camera viewpoint. Temporal video alignment plays an important role in many fine-grained human activity understanding applications, including anomaly detection~\cite{dwibedi2019tcc} (i.e., detect deviations from a dataset of normal videos) and skill transfer~\cite{kwon2022context} (i.e., provide feedback to learners based on expert demonstration videos).

% Self-supervised, RGB-based, 3D skeleton-based approaches, and drawbacks
We are interested in self-supervised approaches for temporal video alignment, since acquiring fine-grained correspondence labels is generally hard and costly~\cite{dwibedi2019tcc,khan2022timestamp}. Early methods~\cite{sermanet2018tcn,dwibedi2019tcc,haresh2021learning,liu2022learning} attempt to match video frames directly. They optimize different temporal alignment metrics such as dynamic time warping~\cite{cuturi2017soft} and optimal transport~\cite{cuturi2013sinkhorn} for self-supervised learning. Although the learned representations show promising generalization to downstream tasks, the results are far from being practical. Recently, CASA~\cite{kwon2022context} proposes a context-aware self-supervised method for temporal video alignment, which operates on 3D skeletons (see Fig.~\ref{fig:teaser}(a)). It conducts 3D skeleton-based augmentations for self-supervised learning, while employing a transformer~\cite{vaswani2017attention} to learn contextual features, yielding great results. Since CASA~\cite{kwon2022context} directly processes 3D skeleton coordinates, its performance is greatly affected by human pose estimation errors such as noisy and missing keypoints.

% Our approach
We present a self-supervised temporal video alignment method, which relies on 2D skeletons (see Fig.~\ref{fig:teaser}(b)). Motivated by the recent successes of~\cite{duan2022revisiting,hyder2024action}, we convert sequences of 2D skeletons into sequences of heatmaps, where each joint is modeled as a Gaussian distribution, making it less sensitive to human pose estimation errors. Unlike CASA~\cite{kwon2022context} which performs self-attention in the temporal domain only, we pass the heatmap sequences to our alignment network, which adopts a video transformer~\cite{arnab2021vivit} to perform self-attention both in the spatial and temporal domains, yielding effective spatiotemporal and contextual features. Moreover, we propose simple various heatmap augmentation strategies based on 2D skeletons for self-supervised learning. As compared to CASA~\cite{kwon2022context}, our method based on 2D skeleton heatmaps obtains both higher accuracy and better robustness against missing and noisy keypoints despite lacking 3D information. Moreover, we show superior performance of our approach over previous methods in fine-grained human activity understanding tasks on three public datasets. Finally, utilizing both 2D skeleton heatmaps and RGB videos as inputs achieves the state-of-the-art across all metrics and datasets.

In summary, our contributions include:
\begin{itemize}
    \item We propose a novel 2D skeleton-based self-supervised temporal video alignment approach for fine-grained human activity understanding. We utilize 2D skeleton heatmaps as input and employ a video transformer to perform self-attention both in the spatial and temporal domains. For self-supervised learning, we introduce simple 2D skeleton heatmap augmentation techniques. 
    \item Our 2D skeleton-based approach is more accurate and robust against noisy and missing keypoints than CASA~\cite{kwon2022context} which requires 3D skeletons. Our evaluations on Penn Action, IKEA ASM, and H2O show that our approach outperforms previous methods on fine-grained human activity understanding tasks. 
    \item We develop a multi-modality variant, achieving the state-of-the-art on all metrics and datasets. To our best knowledge, our work is the first to exploit 2D skeleton heatmap inputs and perform multi-modality fusion for temporal video alignment.
\end{itemize}

%% file: Sections/relatedwork.tex
\section{Related Work}
\label{sec:relatedwork}

\noindent \textbf{Self-Supervised Learning.}
Considerable efforts have been made in developing pretext tasks with pseudo labels for training image-based self-supervised models. Examples include image colorization~\cite{larsson2016learning,larsson2017colorization}, object counting~\cite{noroozi2017representation,liu2018leveraging}, predicting rotations~\cite{gidaris2018unsupervised}, solving puzzles~\cite{carlucci2019domain,kim2019self}, image inpainting~\cite{jenni2020steering}, and image clustering~\cite{caron2018deep,caron2019unsupervised}. Recently, several works~\cite{caron2020unsupervised,chen2020simple,grill2020bootstrap,chen2021exploring} focus on designing data augmentation and contrastive learning strategies, which are important for learning effective representations. Unlike the above image-based methods, video-based methods, which exploit both spatial cues and temporal cues in videos, have attracted a great amount of interests recently. They learn representations by using pretext tasks such as future frame prediction~\cite{srivastava2015unsupervised,vondrick2016generating,ahsan2018discrimnet,diba2019dynamonet} and frame clustering~\cite{kumar2022unsupervised,tran2024permutation}, or leveraging temporal information such as temporal coherence~\cite{mobahi2009deep,zou2011unsupervised,goroshin2015unsupervised}, temporal order~\cite{misra2016shuffle,lee2017unsupervised,fernando2017self,xu2019self}, arrow of time~\cite{pickup2014seeing,wei2018learning}, and pace~\cite{benaim2020speednet,wang2020self,yao2020video}, or utilizing contrastive learning techniques~\cite{feichtenhofer2021large,hu2021contrast,qian2021spatiotemporal,dave2022tclr}.

More recently, skeleton-based methods have been introduced. They learn representations by using pretext tasks such as skeleton inpainting~\cite{zheng2018unsupervised} and motion prediction~\cite{su2020predict}, or utilizing neighborhood consistency~\cite{si2020adversarial}, motion continuity~\cite{su2021self}, and multiple pretext tasks~\cite{lin2020ms2l}. The above methods often do not exploit spatiotemporal dependencies. Here, we use 2D skeleton-based alignment as a pretext task and 2D skeleton-based augmentation for self-supervised learning.

\noindent \textbf{Temporal Video Alignment.}
Despite significant progress in unsupervised time series alignment, methods for self-supervised temporal video alignment have emerged only recently. To learn  representations, TCN~\cite{sermanet2018tcn} employs contrastive learning on synchronized frames across viewpoints, while TCC~\cite{dwibedi2019tcc} imposes cycle-consistent frame correspondences across videos. Recently, LAV~\cite{haresh2021learning} and VAVA~\cite{liu2022learning} adopt dynamic time warping~\cite{cuturi2017soft} and optimal transport~\cite{cuturi2013sinkhorn} respectively as temporal video alignment losses. The above methods focus on aligning video frames directly. More recently, CASA~\cite{kwon2022context} presents a self-supervised temporal video alignment method which operates on 3D skeletons. In this work, we propose a 2D skeleton-based self-supervised temporal video alignment method, which is more accurate and robust against noisy and missing keypoints. We further fuse 2D skeleton heatmaps with RGB videos, establishing the state-of-the-art performance on all metrics and datasets.

\noindent \textbf{Skeleton-Based Action Recognition.}
There exist notable interests in skeleton-based action recognition due to its compactness. A popular group of methods~\cite{yan2018spatial,song2020stronger,cai2021jolo,chen2021channel,gupta2021quo,song2022constructing} model the skeleton sequence as a spatiotemporal graph and rely on Graph Convolutional Networks (GCNs). Despite great results, their drawbacks lie in robustness to noisy/incomplete skeletons~\cite{zhu2019robust}, scalability with multi-person actions, and fusion with other modalities~\cite{das2020vpn}. Another group of methods convert the skeleton sequence into a 2D input (e.g., heatmap/coordinate aggregation~\cite{ke2017new,choutas2018potion,li2018co,luvizon20182d,yan2019pa3d,caetano2019skelemotion,asghari2020dynamic}) or a 3D input (e.g., stacked pseudo images~\cite{hernandez20173d,lin2020image}) and employ Convolutional Neural Networks (CNNs). These methods often suffer from information loss during aggregation. To address that, Duan et al.~\cite{duan2022revisiting} convert each skeleton into a heatmap and pass the heatmap sequence, which preserves all information, to a CNN, and achieve superior results while alleviating the above issues of GCN-based methods. Inspired by their method for \emph{coarse-grained} action recognition, we propose a self-supervised temporal video alignment framework for \emph{fine-grained} tasks, e.g., action phase classification. In addition to video-level cues, fine-grained tasks further require frame-level cues and architectures/losses for extracting/utilizing them, which makes them more challenging.

%% file: Sections/method.tex
\section{Our Approach}
\label{sec:method}

\begin{figure}[t]
	\centering
		\includegraphics[width=1.0\linewidth, trim = 0mm 50mm 0mm 0mm, clip]{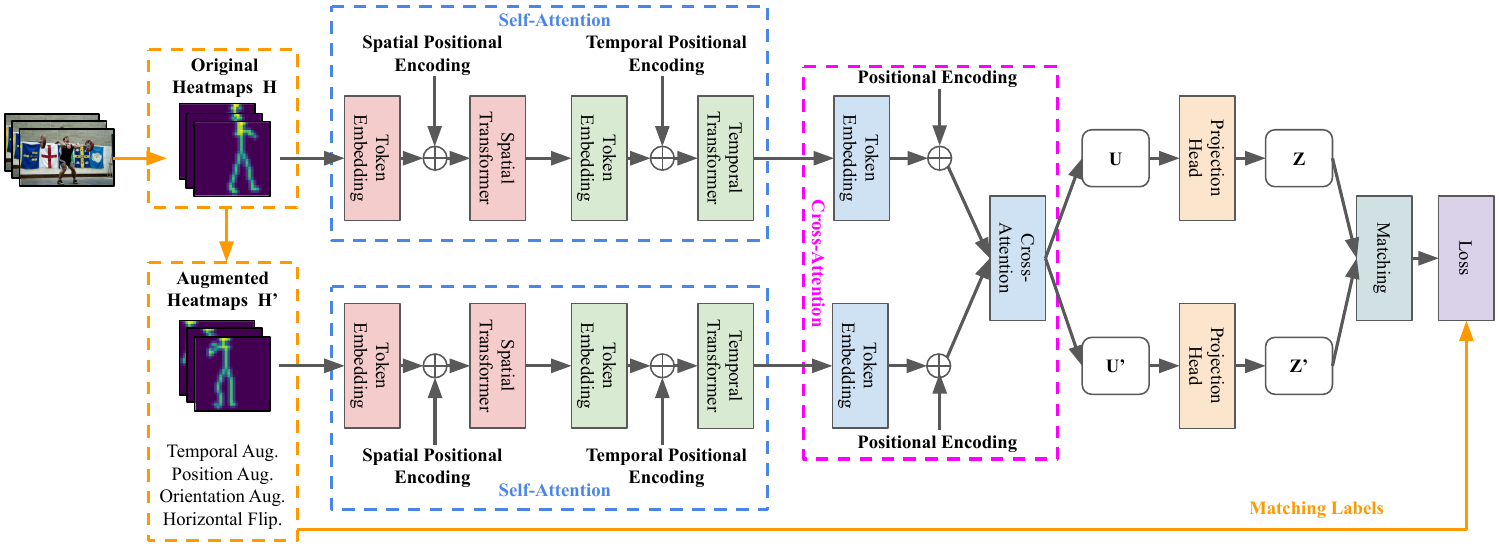}
	\caption{During training, our approach takes as input sequences of original heatmaps and augmented heatmaps. We perform self-attention both in the spatial and temporal domains to extract effective spatiotemporal and contextual cues within each sequence and cross-attention to extract contextual cues across the sequences. The extracted features after projection heads are passed to a matching module to predict correspondences across the sequences. Matching labels generated by the augmentation module are used in the loss function.}
	\label{fig:method}
\end{figure}

Below we describe our main contribution, a self-supervised learning method using 2D skeleton-based temporal video alignment as a pretext task. Fig.~\ref{fig:method} illustrates an overview. 

\noindent \textbf{Notations.} Let us denote $\boldsymbol{S} = \{\boldsymbol{s}_1, \boldsymbol{s}_2, \dots, \boldsymbol{s}_M\}$ and $\boldsymbol{S}' = \{\boldsymbol{s}'_1, \boldsymbol{s}'_2, \dots, \boldsymbol{s}'_N\}$ as the original 2D skeleton sequence and its augmented sequence, with length $M$ and $N$ respectively. Each 2D skeleton has $K$ 2D joints, i.e., $\boldsymbol{s}_i \in \mathbb{R}^{K \times 2}$ and $\boldsymbol{s}'_j \in \mathbb{R}^{K \times 2}$. The heatmaps obtained from $\boldsymbol{S}$ and $\boldsymbol{S}'$ are written as $\boldsymbol{H} = \{\boldsymbol{h}_1, \boldsymbol{h}_2, \dots, \boldsymbol{h}_M\}$ and $\boldsymbol{H}' = \{\boldsymbol{h}'_1, \boldsymbol{h}'_2, \dots, \boldsymbol{h}'_N\}$, with $\boldsymbol{h}_i$ and $\boldsymbol{h}'_j$ computed from $\boldsymbol{s}_i$ and $\boldsymbol{s}'_j$ respectively. Next, let us represent the embedding function (up to the cross attention module) as $f_{\boldsymbol{\theta}}$, with learnable parameters $\boldsymbol{\theta}$. The embedding features of $\boldsymbol{H}$ and $\boldsymbol{H}'$ are expressed as $\boldsymbol{U} = \{\boldsymbol{u}_1, \boldsymbol{u}_2, \dots, \boldsymbol{u}_M\}$ and $\boldsymbol{U}' = \{\boldsymbol{u}'_1, \boldsymbol{u}'_2, \dots, \boldsymbol{u}'_N\}$, with $\boldsymbol{u}_i = f_{\boldsymbol{\theta}}(\boldsymbol{h}_i)$ and $\boldsymbol{u}'_j = f_{\boldsymbol{\theta}}(\boldsymbol{h}'_j)$ respectively. Lastly, we denote the projection head function as $g_{\boldsymbol{\phi}}$, with learnable parameters $\boldsymbol{\phi}$, and the latent features before the loss function as $\boldsymbol{Z} = \{\boldsymbol{z}_1, \boldsymbol{z}_2, \dots, \boldsymbol{z}_M\}$ and $\boldsymbol{Z}' = \{\boldsymbol{z}'_1, \boldsymbol{z}'_2, \dots, \boldsymbol{z}'_N\}$, with $\boldsymbol{z}_i = g_{\boldsymbol{\phi}}(\boldsymbol{u}_i)$ and $\boldsymbol{z}'_j = g_{\boldsymbol{\phi}}(\boldsymbol{u}'_j)$ respectively.

\subsection{2D Skeleton Heatmap}
\label{sec:heatmap}

We discuss the estimation of 2D skeletons $\boldsymbol{S}$ and their corresponding heatmaps $\boldsymbol{H}$ from a video sequence. An off-the-shelf 2D human pose estimator such as OpenPose~\cite{cao2019openpose} can be used to extract good quality 2D skeletons. For memory efficiency, a 2D skeleton is often stored as a set of 2D joint triplets $\{(x_k,y_k,c_k)\}$, where $(x_k,y_k)$ are the 2D coordinates and $c_k$ is the (maximum) confidence score of the $k$-th joint. In practice, for fair comparisons, we use 2D skeletons that are projected (via orthographic projection) from 3D skeletons provided by CASA~\cite{kwon2022context} and set $c_k = 1.0$. We use orthographic projection since camera intrinsics/extrinsics are not available for all datasets (e.g., Penn Action). Our results with OpenPose 2D poses can be found in the supplementary material.

We now convert 2D skeletons into heatmaps. For a 2D skeleton, we compute a heatmap of size $K \times H \times W$, where $K$ is the number of 2D joints, and $H$ and $W$ are the height and width of the video frame respectively. Given a set of 2D joint triplets $\{(x_k,y_k,c_k)\}$, a \emph{joint} heatmap $\boldsymbol{h}^{J}$ consisting of $K$ Gaussian maps centered at every joint is derived as:
\begin{align}
    \boldsymbol{h}^{J}_{kij} = e^{-\frac{(i-x_k)^2+(j-y_k)^2}{2*\sigma^2}} * c_k,
\end{align}
where $\sigma$ is the standard deviation of the Gaussian maps. Alternatively, a \emph{limb} heatmap $\boldsymbol{h}^{L}$ of size $L \times H \times W$ ($L$ is the number of limbs in a skeleton) is written as: 
\begin{align}
    \boldsymbol{h}^{L}_{lij} = e^{-\frac{dist((i,j), seg(a_l,b_l))}{2*\sigma^2}} * min(c_{a_l},c_{b_l}),
\end{align}
where the $l$-th limb is the segment $seg(a_l,b_l)$ between the two joints $a_l$ and $b_l$, and the $dist$ function computes the distance from the location $(i,j)$ to the segment $seg(a_l,b_l)$. As we will show later in Sec.~\ref{sec:ablation_heatmaps}, combining both joint and limb heatmaps as input to our alignment network yields the best results, i.e., for each 2D skeleton $\boldsymbol{s}_i$, we extract the combined heatmap $\boldsymbol{h}_i = \boldsymbol{h}^{J+L}_i$. Moreover, since the action of interest appears only in a small region in the video frames, we crop the heatmaps along the spatial dimensions using the smallest bounding box containing all 2D skeletons in the video sequence and resize them to the spatial dimensions of $30 \times 30$. Thus, we reduce the size of the heatmaps, while preserving all 2D skeletons and their motions.

\noindent \textbf{2D Skeleton Heatmaps vs. 3D Skeleton Coordinates.} Unlike CASA~\cite{kwon2022context} which operates on 3D skeleton coordinates, our method relies on 2D skeleton heatmaps. Advantages of 2D skeleton heatmaps over 3D skeleton coordinates include: 1) Accurate 2D skeletons are generally easier to obtain, as accurate depths are not required (e.g., see Fig.~1 of [22]), 2) 2D skeleton heatmaps are more robust, as joints/limbs are modeled by Gaussian maps, 3) 2D skeleton heatmaps are applicable to advanced CNNs, whereas MLPs/GCNs are often used for 3D skeleton coordinates, and 4) It is convenient to fuse 2D skeleton heatmaps with other grid-like data. The above advantages contribute to our superior accuracy and robustness, as shown later in Sec.~\ref{sec:experiments}.

\subsection{2D Skeleton Heatmap Augmentation}
\label{sec:augmentation}

\begin{figure}[t]
	\centering
		\includegraphics[width=0.4\linewidth, trim = 0mm 50mm 135mm 0mm, clip]{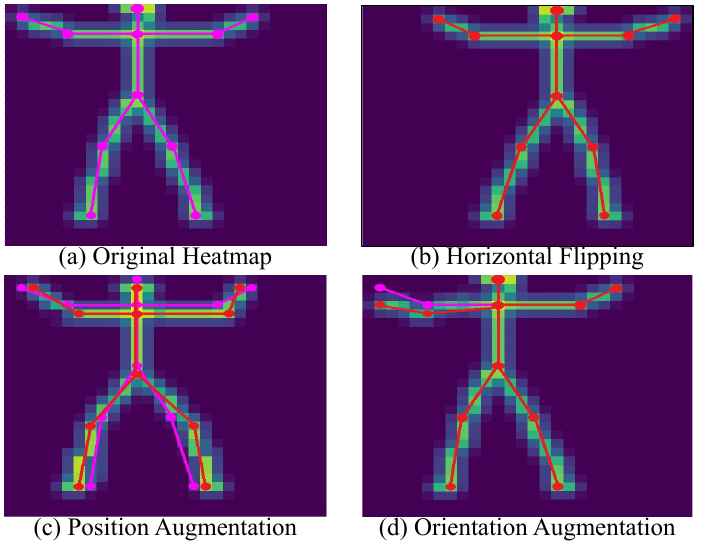}
	\caption{Examples of our heatmap augmentation techniques based on 2D skeletons. The original 2D skeleton is in pink, while the augmented 2D skeleton is in red.}
	\label{fig:augmentation}
\end{figure}

Below we describe our simple heatmap augmentation techniques based on 2D skeletons, which are used to generate correspondence labels for self-supervised training of our alignment network. This is in contrast with the coordinate augmentation techniques in CASA~\cite{kwon2022context}, which rely on 3D skeletons. We first apply various augmentations on the original 2D skeletons $\boldsymbol{S}$ to obtain the augmented 2D skeletons $\boldsymbol{S}'$, and then compute the augmented heatmaps $\boldsymbol{H}'$ from the augmented 2D skeletons $\boldsymbol{S}'$. They include:

\noindent \textbf{Temporal Augmentation.} We adjust action speed by randomly selecting $N$ 2D skeletons out of $M$ 2D skeletons in $\boldsymbol{S}$, resulting in $\boldsymbol{S}'$.

\noindent \textbf{Position Augmentation.} We perturb 2D joints in $\boldsymbol{S}$ by adding a random noise (sampled from a Gaussian distribution $\mathcal{N}(\nu)$ with a standard deviation $\nu$), yielding $\boldsymbol{S}'$.

\noindent \textbf{Orientation Augmentation.} We pick a limb in $\boldsymbol{S}$ (arm or leg) and rotate it by a random angle (sampled from a Gaussian distribution $\mathcal{N}(\rho)$ with a standard deviation $\rho$), resulting in $\boldsymbol{S}'$.

\noindent \textbf{Horizontal Flipping.} We flip 2D joints in $\boldsymbol{S}$ (left to right or vice versa), yielding $\boldsymbol{S}'$.

Fig.~\ref{fig:augmentation} shows examples of our simple heatmap augmentation techniques based on 2D skeletons. For position and orientation augmentations, we empirically use a multivariate Gaussian distribution with a covariance matrix so that the added noise is temporally smooth as in CASA~\cite{kwon2022context}. However, it does not lead to improved results and is thus not used in the above. Furthermore, the above augmentations generate correspondence labels between $\boldsymbol{S}$ and $\boldsymbol{S}'$ (and similarly, between $\boldsymbol{H}$ and $\boldsymbol{H}'$), which can be used for self-supervised training of our alignment network in the next section.

\subsection{2D Skeleton-Based Self-Supervised Temporal Video Alignment}
\label{sec:alignment}

\begin{figure}[t]
	\centering
		\includegraphics[width=0.5\linewidth, trim = 0mm 85mm 140mm 0mm, clip]{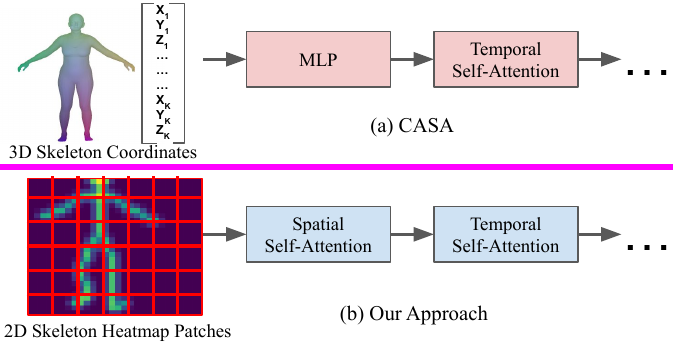}
	\caption{(a) CASA directly processes 3D skeleton coordinates and conducts self-attention in the temporal domain only. (b) Our approach operates on 2D skeleton heatmaps and performs self-attention both in the spatial and temporal domains.}
	\label{fig:selfattetion}
\end{figure}

Below we describe our network architecture and loss:

\noindent \textbf{Self-Attention.} Given the original heatmap sequence $\boldsymbol{H}$ and the augmented sequence $\boldsymbol{H}'$, we pass them separately through a video transformer, namely ViViT~\cite{arnab2021vivit}, to perform self-attention both in the spatial and temporal domains, yielding effective spatiotemporal and contextual cues within each sequence. Below we discuss the processing of $\boldsymbol{H}$ only since the same is applied for $\boldsymbol{H}'$. For each $\boldsymbol{h}_i$ in $\boldsymbol{H}$, we first break it into $h * w$ non-overlapping patches, which are embedded via linear projection into $h * w$ tokens. The tokens from $\boldsymbol{h}_i$ (with spatial positional encoding) are then fed to a spatial transformer, which learns interactions between tokens from the same temporal index (i.e., between joints in the same skeleton). Next, we employ global average pooling on the above tokens output by the spatial transformer to obtain the frame-level token representing $\boldsymbol{h}_i$. The frame-level tokens of $\boldsymbol{H}$ (with temporal positional encoding) are passed together to a temporal transformer, which learns interactions between tokens from different temporal indexes (i.e., between skeletons in the same sequence). Note that our approach performs self-attention both in the spatial and temporal domains, whereas CASA~\cite{kwon2022context} conducts self-attention in the temporal domain only. Fig.~\ref{fig:selfattetion} illustrates this difference. In addition, our positional encodings are learned, whereas it is fixed in CASA~\cite{kwon2022context}. 

\noindent \textbf{Cross-Attention.} The above modules only focus on self-attention in the spatial and temporal domains within each sequence. Following~\cite{sun2021loftr,kwon2022context}, we add a cross-attention module to learn interactions between $\boldsymbol{H}$ and $\boldsymbol{H}'$ (i.e., between skeletons across the sequences), yielding contextual cues across the sequences. In particular, we encode the tokens output by the self-attention modules via an MLP layer and positional encoding before applying cross-attention.

\noindent \textbf{Projection Head.} We include a projection head, which is an MLP network with one hidden layer. As mentioned in~\cite{chen2020simple,kwon2022context}, adding the projection head improves the generalization ability and yields effective features for downstream tasks.

\noindent \textbf{Matching and Loss.} We first compute the probability $\boldsymbol{\omega}_{ij}$ that the $i$-th frame in $\boldsymbol{H}$ is matched with the $j$-th frame in $\boldsymbol{H}'$ using contrastive regression~\cite{dwibedi2019tcc,kwon2022context} as:
\begin{align}
    \boldsymbol{\omega}_{ij} = \frac{e^{-|| \boldsymbol{z}_i - \boldsymbol{z}'_j||/\lambda}}{\sum^M_{m=1} e^{-|| \boldsymbol{z}_m - \boldsymbol{z}'_j||/\lambda}},
\end{align}
 where $\lambda$ is a temperature parameter. The above formulation encourages temporally close frames to be mapped to nearby points and temporally distant frames to be mapped to far away points in the latent space. The frame index $i^{pred}_j$ in $\boldsymbol{H}$ corresponding to the $j$-th frame in $\boldsymbol{H}'$ is then predicted as $i^{pred}_j = \sum^M_{i=1} i*\boldsymbol{\omega}_{ij}$. The matching labels between $\boldsymbol{H}$ and $\boldsymbol{H}'$ are obtained in the augmentation module and hence can be used as supervision signals for training our alignment network. The difference between the predicted frame index $i^{pred}_j$ and ground truth frame index $i^{gt}_j$ is minimized as:
\begin{align}
    \mathcal{L} = \frac{1}{N} \sum^N_{j=1} || i^{pred}_j - i^{gt}_j ||^2.
\end{align}

\subsection{Multi-Modality Fusion}
\label{sec:fusion}

Representing skeletons as heatmaps makes it convenient to fuse with other modalities with grid structures such as RGB videos, optical flows, and depth maps. In this section, we extend our approach by using both skeleton heatmaps and RGB videos as inputs and performing late fusion. Due to space constraints, an illustration of our RGB+Pose model architecture is included in the supplementary material. Our RGB+Pose model begins with separate encoders which respectively map the original skeleton heatmaps $\boldsymbol{H}$ and the augmented skeleton heatmaps $\boldsymbol{H}'$ to features $\boldsymbol{U_H}$ and $\boldsymbol{U_H}'$, and respectively map the original video frames $\boldsymbol{F}$ and the augmented video frames $\boldsymbol{F}'$ to features $\boldsymbol{U_F}$ and $\boldsymbol{U_F}'$. For heatmap augmentation, we apply the techniques in Sec.~\ref{sec:augmentation}, while for video frame augmentation, we employ temporal augmentation, brightness augmentation, contrast augmentation, and horizontal flipping. Next, $\boldsymbol{U_H}$ and $\boldsymbol{U_F}$ are fused together to yield the combined features $\boldsymbol{U}$ for the original sequences, and $\boldsymbol{U_H}'$ and $\boldsymbol{U_F}'$ are fused together to yield the combined features $\boldsymbol{U}'$ for the augmented sequences. Our fusion module includes concatenating the feature vectors and a linear layer to reduce the concatenated feature vector to the original size of 192. The fused features $\boldsymbol{U}$ and $\boldsymbol{U}'$ are then fed to the projection heads to obtain the final features $\boldsymbol{Z}$ and $\boldsymbol{Z}'$, which are used to perform matching and compute the loss as in Fig.~\ref{fig:method}. To train our RGB+Pose model, we perform two-stage training, i.e., we first train the encoders for each modality separately using the single-modality alignment loss, and then train them jointly with late fusion.

\noindent \textbf{Differences with CASA~\cite{kwon2022context}.} There are several differences between our method and CASA~\cite{kwon2022context}, including: 1) Our method simply requires 2D skeletons (vs. 3D skeletons in CASA). 2) Our method performs self-attention both in spatial and temporal domains (vs. no spatial self-attention in CASA). 3) Our simple heatmap augmentations are based on 2D skeletons (vs. 3D skeletons in CASA). 4) Our method obtains both higher accuracy and better robustness than CASA. 5) Our multi-modality fusion further improves the performance (vs. no multi-modality fusion in CASA).

%% file: Sections/experiments.tex
\section{Experiments}
\label{sec:experiments}

\begin{figure}[t]
	\centering
		\includegraphics[width=1.0\linewidth, trim = 0mm 85mm 0mm 0mm, clip]{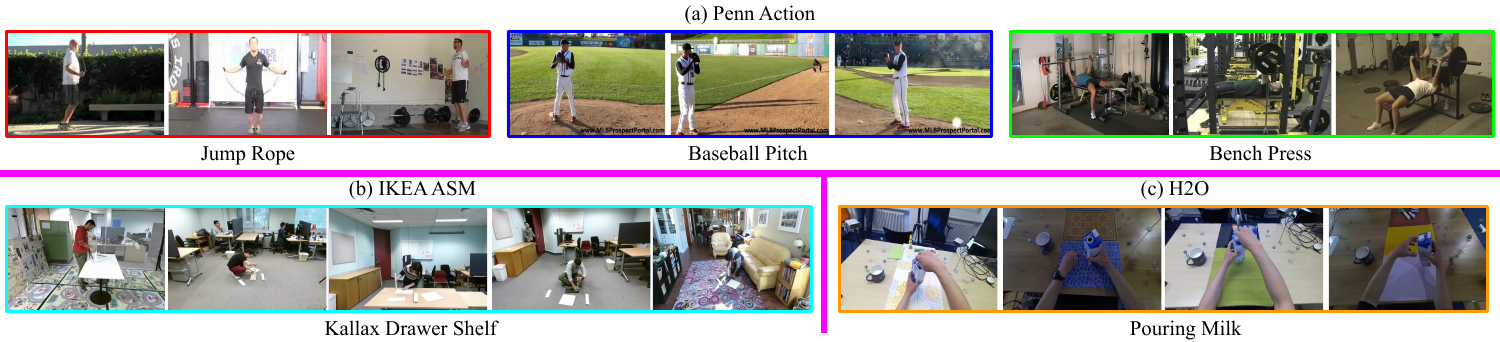}
	\caption{Example videos with diverse camera viewpoints, actors, backgrounds, and actions from Penn Action, IKEA ASM, and H2O.}
	\label{fig:diverse_viewpoints}
\end{figure}

\noindent \textbf{Datasets.} We evaluate on IKEA ASM~\cite{ikeaasm}, H2O~\cite{h2odataset}, and Penn Action~\cite{pennaction}. Following~\cite{haresh2021learning,liu2022learning,kwon2022context}, we experiment on the \textit{Kallax Drawer Shelf} videos of IKEA ASM~\cite{ikeaasm}, showing a furniture assembly activity. H2O~\cite{h2odataset} is a hand-object interaction dataset. We evaluate on the \textit{Pouring Milk} activity of H2O, following~\cite{kwon2022context}. Penn Action~\cite{pennaction} contains sporting activities. We use the same 13 actions of Penn Action as~\cite{dwibedi2019tcc,haresh2021learning,liu2022learning,kwon2022context}. These datasets are captured with diverse camera viewpoints, actors, backgrounds, and actions, as illustrated in Fig.~\ref{fig:diverse_viewpoints}. We use full-body poses for Penn Action~\cite{pennaction} and IKEA ASM~\cite{ikeaasm}, hand poses for H2O~\cite{h2odataset}, and the same training/testing splits as~\cite{dwibedi2019tcc,haresh2021learning,liu2022learning,kwon2022context}.

\noindent \textbf{Implementation Details.}
Our approach is implemented in pyTorch~\cite{paszke2017automatic}. We randomly initialize our model and use ADAM~\cite{kingma2014adam} optimization. For Penn Action~\cite{pennaction}, one model is trained separately for each of the 13 actions. For datasets with a single action such as IKEA ASM~\cite{ikeaasm} and H2O~\cite{h2odataset}, a single model is trained. Due to space limits, more details are available in the supplementary material.

\noindent \textbf{Competing Methods.}
We test our approach, namely \emph{LA2DS} (short for \emph{L}earning by \emph{A}ligning \emph{2D} \emph{S}keletons), against state-of-the-art self-supervised temporal video alignment methods, including SAL~\cite{ishan2016sal}, TCN~\cite{sermanet2018tcn}, TCC~\cite{dwibedi2019tcc}, LAV~\cite{haresh2021learning}, VAVA~\cite{liu2022learning}, and CASA~\cite{kwon2022context}. Moreover, we combine CASA~\cite{kwon2022context} with the graph convolutional network STGCN~\cite{yan2018spatial}, and integrate LAV~\cite{haresh2021learning} with the video transformer ViViT~\cite{arnab2021vivit} (pre-trained on Kinetics-400~\cite{kinetics400}) as additional competing methods.

\noindent \textbf{Evaluation Metrics.} We compute four metrics on the validation set, i.e., Phase Classification Accuracy, Phase Progression, Kendall’s Tau, and Average Precision. Please refer to~\cite{dwibedi2019tcc,haresh2021learning,liu2022learning,kwon2022context} for the definitions. We train the network on the training set and freeze it, and then train an SVM classifier or a linear regressor on top of the frozen features.

\subsection{Ablation Results}
\label{sec:exp_ablation}

In the following, we perform ablation experiments on the H2O dataset to evaluate the effectiveness of our design choices in Sec.~\ref{sec:method}. Moreover, we study the robustness of our approach in the presence of missing and noisy keypoints.

\subsubsection{Effects of Different Network Components}
\input{Tables/ablation_table_components.tex}
The ablation results of various network components in Tab.~\ref{tab:ablation_components} demonstrate that all of our network components in Sec.~\ref{sec:alignment} consistently enhance the performance, and removing any of them leads to a performance drop. For example, removing spatial and temporal transformers causes a performance decline because self-attention captures spatiotemporal and contextual cues within a sequence. More importantly, removing spatial transformer yields the worst results on both phase progression and Kendall's Tau.

\subsubsection{Effects of Different Heatmaps}
\label{sec:ablation_heatmaps}
\input{Tables/ablation_table_heatmaps.tex}
We now conduct an ablation study by using various heatmaps in Sec.~\ref{sec:heatmap} as input to our network, including joint heatmaps only, limb heatmaps only, and combined joint+limb heatmaps. Tab.~\ref{tab:ablation_heatmaps} demonstrates the results. It is evident from Tab.~\ref{tab:ablation_heatmaps} that using integrated joint+limb heatmaps as input to our network leads to the best results, outperforming using joint heatmaps or limb heatmaps alone.

\subsubsection{Robustness against Missing Keypoints}
\input{Tables/ablation_keypoint_missing.tex}
To assess the robustness of our method, we perform an experiment where a certain part of the skeleton is removed. Specifically, we randomly remove one finger with a probability $p$ for each frame in the H2O dataset during testing (without any finetuning) and measure the impact of this perturbation on the classification accuracy. The results in Tab.~\ref{tab:ablation_keypoint_missing} show that LA2DS exhibits a high level of robustness to missing keypoints. For example, when one finger is dropped per frame (i.e., $p = 100\%$), the decrease in accuracy for LA2DS is minimal (about $1\%$), whereas for CASA~\cite{kwon2022context}, the drop is around $8\%$. This is likely because our method represents each joint by a Gaussian distribution and performs self-attention in the spatial domain to leverage dependencies between joints in the same skeleton, whereas CASA models each joint merely by its coordinates and does not conduct spatial self-attention.

\subsubsection{Robustness against Noisy Keypoints}
\input{Tables/ablation_keypoint_quality.tex}
To examine the generalization of our approach to noisy keypoints, we train and test on skeletons of different qualities. In particular, we use ground truth keypoints provided by the H2O dataset, namely \emph{clean} keypoints, as well as add random noise to them, yielding \emph{noisy} keypoints. We measure the impact of this perturbation on the classification accuracy. The results in Tab.~\ref{tab:ablation_keypoint_quality} show that LA2DS exhibits a high level of robustness against noisy keypoints. For instance, when noisy keypoints are used for both training and testing, the accuracy of CASA~\cite{kwon2022context} is decreased by more than 5\%, whereas our method has a drop of less than 2\%. Similar to the prior section, our method is more robust to noisy keypoints than CASA, thanks to our heatmap representation and self-attention in the spatial domain.

\subsection{Comparison Results} 
\label{sec:exp_comparison}

Below we quantitatively compare our approach with previous methods. In addition, some qualitative comparisons are provided in the supplementary material.

\subsubsection{Phase Classification Results}
\input{Tables/phase_classification.tex}
We first evaluate the effectiveness of our learned representations for action phase classification. Tab.~\ref{tab:phase_classification} reports the quantitative results of all methods on Penn Action, IKEA ASM, and H2O. It is evident that our method achieves the best overall performance, despite diverse camera viewpoints, actors, backgrounds, and actions (e.g., see Fig.~\ref{fig:diverse_viewpoints}). The inferior performance of ViViT~\cite{arnab2021vivit}+LAV~\cite{haresh2021learning} is likely because ViViT~\cite{arnab2021vivit} has roughly one fifth number of parameters of ResNet-50~\cite{he2016deep} used in the original LAV~\cite{haresh2021learning}. In contrast, we empirically observe that STGCN~\cite{yan2018spatial}+CASA~\cite{kwon2022context} often suffers from overfitting due to low-dimensional input vectors. Also, we notice that RGB-based methods (i.e., LAV, VAVA) often outperform both 2D and 3D skeleton-based methods (i.e., CASA, LA2DS) in limited label settings on IKEA ASM, since different actions may have similar skeletons on IKEA ASM (e.g., picking up front panel, picking up back panel) and context details from RGB data are important for classifying those actions. We will later address this limitation by multi-modality fusion (see Sec.~\ref{sec:exp_fusion}). The results in Tab.~\ref{tab:phase_classification} indicate that LA2DS is effective in learning spatiotemporal and contextual features, yielding superior classification performance. 

\subsubsection{Phase Progression and Kendall's Tau Results}
\input{Tables/progress_tau.tex}
Tab.~\ref{tab:progress_tau} presents the phase progression and Kendall's Tau results on Penn Action and H2O (IKEA ASM is not included due to its repetitive labels~\cite{haresh2021learning}). Phase progression measures how well an action progresses over the video, whereas Kendall's Tau measures how well two sequences are aligned temporally. From Tab.~\ref{tab:progress_tau}, it can be seen that our approach obtains the best performance in Kendall's Tau, while having comparably best numbers along with CASA~\cite{kwon2022context} in phase progression.  This suggests that our method is capable of capturing dependencies between skeletons across sequences, which play a crucial role for temporal alignment.

\subsubsection{Fine-Grained Frame Retrieval Results}
\input{Tables/frame_retrieval.tex}
Here, we examine the performance of our learned representations for fine-grained frame retrieval on Penn Action, IKEA ASM, and H2O, and compare it with previous methods as illustrated in Tab.~\ref{tab:frame_retrieval}. Following~\cite{haresh2021learning}, our experimental setup involves using one video from the validation set as the query video, while the remaining videos in the validation set serve as the support set. To retrieve $K$ most similar frames in the support set for each query frame in the query video, we find $K$ nearest neighbors in the embedding space. We then calculate Average Precision at $K$, which represents the percentage of the $K$ retrieved frames with the same action phase label as the query frame. It is obvious from Tab.~\ref{tab:frame_retrieval} that our method achieves the best results, outperforming all competing methods across all datasets and all metrics, despite diverse camera viewpoints, actors, backgrounds, and actions (e.g., see Fig.~\ref{fig:diverse_viewpoints}). The results in Tab.~\ref{tab:frame_retrieval} validate the effectiveness of our learned representations, which capture dependencies between joints in the same skeleton as well as dependencies between skeletons in the same sequence, for fine-grained frame retrieval.

\subsection{Fusion Results}
\label{sec:exp_fusion}

\input{Tables/fusion.tex}
Tab.~\ref{tab:fusion} presents the multi-modality fusion results on Penn Action only. The complete results on all metrics and datasets are included in the supplementary material. For comparison purposes, we extend CASA~\cite{kwon2022context} by adding branches with RGB inputs and ViViT encoders and performing RGB+3D skeleton fusion via late fusion (similar to our RGB+2D skeleton fusion in Sec.~\ref{sec:fusion}), yielding the CASA-based fusion baseline. It is evident from Tab.~\ref{tab:fusion} that our fusion model achieves the best results, outperforming all the single-modality methods and the CASA-based fusion baseline and establishing the state-of-the-art. Both our fusion model and the CASA-based fusion baseline outperform their single-modality counterparts respectively, which confirms the benefit of multi-modality inputs for fine-grained human activity understanding tasks. 

\subsection{Comparisons with Recent RGB-Based Methods}
\label{sec:recentmethods}

We compare our approach with recent RGB-based methods, namely VSP~\cite{zhang2023modeling} (video inputs) and STEPs~\cite{shah2023steps} (video and optical flow inputs). These methods do not employ temporal alignment losses. The results on IKEA ASM are shown in Tab.~\ref{tab:recent_methods}. Our method outperforms VSP and STEPs on very few metrics, i.e., better retrieval results but worse classification results than VSP, and better Acc@0.1 but worse Acc@0.5 and Acc@1.0 than STEPs. Our worse performance is likely due to our compact inputs (e.g., missing context details). Nevertheless, the key advantage of our method is privacy preserving, which is important for several applications involving human action understanding.

\begin{table}[!t]
\scriptsize
\begin{minipage}{\linewidth}
\centering
{%
\setlength{\tabcolsep}{7pt}
\begin{tabular}{c|c|c|c|c|c|c|c|c}
\specialrule{1pt}{1pt}{1pt}
& \textbf{\scriptsize{Method}} & \textbf{\scriptsize{Input}} & \textbf{\scriptsize{Acc@0.1}} & \textbf{\scriptsize{Acc@0.5}} & \textbf{\scriptsize{Acc@1.0}} & \textbf{\scriptsize{AP@5}} & \textbf{\scriptsize{AP@10}} & \textbf{\scriptsize{AP@15}} \\

\midrule
\multirow{4}{*}{\rotatebox[origin=c]{90}{\textbf{\scriptsize{IKEA ASM}}}}
&VSP~\cite{zhang2023modeling} & $\square$ & \textbf{45.00} & \textbf{45.97} & \textbf{47.52} & 30.23 & - & - \\
&STEPs~\cite{shah2023steps} & $\approx$ + $\square$ & 28.59 & \underline{\textit{36.25}} & \underline{\textit{37.02}} & - & - & - \\
&\cellcolor{beaublue}LA2DS (Ours)
&\cellcolor{beaublue}$\star$ &\cellcolor{beaublue}26.43 &\cellcolor{beaublue}32.56 &\cellcolor{beaublue}34.73 &\cellcolor{beaublue}\underline{\textit{32.44}} &\cellcolor{beaublue}\underline{\textit{31.89}} &\cellcolor{beaublue}\underline{\textit{31.56}} \\
&\cellcolor{beaublue}LA2DS (Ours) 
&\cellcolor{beaublue}$\star$ + $\square$  &\cellcolor{beaublue}\underline{\textit{31.74}} &\cellcolor{beaublue}34.51 &\cellcolor{beaublue}36.18 &\cellcolor{beaublue}\textbf{33.08} &\cellcolor{beaublue}\textbf{32.25} &\cellcolor{beaublue}\textbf{32.17} \\

\specialrule{1pt}{1pt}{1pt}
\end{tabular}
}
\caption{Comparisons with recent RGB-based methods. Note that $\square$ denotes direct RGB inputs, $\approx$ denotes optical flow inputs, and $\star$ denotes 2D pose heatmap inputs. Best results are in \textbf{bold}, while second best ones are \underline{\textit{underlined}}.}
\label{tab:recent_methods}
\end{minipage}
\end{table}

\noindent \textbf{Supplementary Material.} We provide additional results and discussions in the supplementary material. They include our results with different method parts, which analyze sources of performance gains, our results with OpenPose~\cite{cao2019openpose} 2D poses, which are better than our above results with projected 2D poses, and discussions of limitations.

%% file: Tables/ablation_table_components.tex
\begin{table}[!t]
\scriptsize
\begin{minipage}[b]{1.0\linewidth}
\centering

{%
\setlength{\tabcolsep}{2pt}
\begin{tabular}{c|c|c|c|c}

\specialrule{1pt}{1pt}{1pt}

 & \textbf{\scriptsize{Method}} & \textbf{\scriptsize{Class.}} & \textbf{\scriptsize{Progress.}} & \bm{$\tau$}\\
\midrule

\multirow{6}{*}{\rotatebox[origin=c]{90}{\textbf{\scriptsize{H2O}}}}
&w/o Temp. Transf.
& 64.40&0.9149&0.9566
\\
&w/o Proj. Head
&  66.53  & 0.9169 & 0.9590
\\
&w/o Pos. Enc.
&  67.80  &  0.9196 & \underline{\textit{0.9615}}
\\
&w/o Cross Attent.
&  67.83&\underline{\textit{0.9198}}&0.9590
  \\
&w/o Spat. Transf.
&  \underline{\textit{68.47}}  & 0.9039 & 0.9548
\\
&All
&\textbf{70.12}    &\textbf{0.9280}  &\textbf{0.9670}
\\

\specialrule{1pt}{1pt}{1pt}
\end{tabular}

}

\caption{Effects of different network components. Best results are in \textbf{bold}, while second best ones are \underline{\textit{underlined}}.}
%\vspace{-0.5cm}
\label{tab:ablation_components}
\end{minipage}

\end{table}

%% file: Tables/ablation_table_heatmaps.tex
\begin{table}[!t]
\scriptsize
\begin{minipage}[b]{1.0\linewidth}
\centering

{%
\setlength{\tabcolsep}{2pt}
\begin{tabular}{c|c|c|c|c}

\specialrule{1pt}{1pt}{1pt}

 & \textbf{\scriptsize{Method}} & \textbf{\scriptsize{Class.}} & \textbf{\scriptsize{Progress.}} & \bm{$\tau$}\\
\midrule

\multirow{3}{*}{\rotatebox[origin=c]{90}{\textbf{\scriptsize{H2O}}}}
&Limbs
& \underline{\textit{67.72}} &0.9099 &0.9614

  \\
 &Joints
&  66.09&\underline{\textit{0.9101}}&\underline{\textit{0.9646}}

\\
&Limbs+Joints
&\textbf{70.12}    &\textbf{0.9280}  &\textbf{0.9670}

\\

\specialrule{1pt}{1pt}{1pt}
\end{tabular}

}

\caption{Effects of different heatmaps. Best results are in \textbf{bold}, while second best ones are \underline{\textit{underlined}}.}
%\vspace{-0.5cm}
\label{tab:ablation_heatmaps}
\end{minipage}

\end{table}

%% file: Tables/ablation_keypoint_missing.tex
\begin{table}[!t]
\scriptsize
\begin{minipage}[b]{1.0\linewidth}
\centering

{%
\setlength{\tabcolsep}{2pt}
\begin{tabular}{c|c|c|c|c|c|c}

\specialrule{1pt}{1pt}{1pt}

 & \textbf{\scriptsize{Method}} & \textbf{\scriptsize{Input}}& \multicolumn{4}{c}{\textbf{\scriptsize{Keypoint Missing Probability $p$}}} \\ 
 \cline{4-7}
&&&\textbf{\scriptsize{0\%}}&\textbf{\scriptsize{25\%}}&\textbf{\scriptsize{50\%}}&\textbf{\scriptsize{100\%}}\\

\midrule
\multirow{2}{*}{\rotatebox[origin=c]{90}{\textbf{\scriptsize{H2O}}}}
&CASA \cite{kwon2022context} &$\blacktriangle$& 68.78  & 66.68 & 64.80 & 60.93 \\
&\cellcolor{beaublue} LA2DS (Ours) &\cellcolor{beaublue}$\star$ &\cellcolor{beaublue} \textbf{70.12}   &\cellcolor{beaublue} \textbf{69.54}  &\cellcolor{beaublue} \textbf{69.02} &\cellcolor{beaublue} \textbf{68.94}   \\
\specialrule{1pt}{1pt}{1pt}
\end{tabular}

}

\caption{Robustness against missing keypoints. Note that $\blacktriangle$ denotes 3D pose coordinate inputs, while $\star$ denotes 2D pose heatmap inputs. Best results are in \textbf{bold}.}
%\vspace{-0.5cm}
\label{tab:ablation_keypoint_missing}
\end{minipage}

\end{table}

%% file: Tables/ablation_keypoint_quality.tex
\begin{table}[!t]
\scriptsize
\begin{minipage}[b]{1.0\linewidth}
\centering

{%
\setlength{\tabcolsep}{2pt}
\begin{tabular}{c|c|c|c|c|c|c}

\specialrule{1pt}{1pt}{1pt}

 & \textbf{\scriptsize{Method}} & \textbf{\scriptsize{Input}}& \multicolumn{4}{c}{\textbf{\scriptsize{Train Data~$\Rightarrow$~Test Data}}} \\ 
 \cline{4-7}
&&&\textbf{\scriptsize{C~$\Rightarrow$~C}}&\textbf{\scriptsize{C~$\Rightarrow$~N}}&\textbf{\scriptsize{N~$\Rightarrow$~C}}&\textbf{\scriptsize{N~$\Rightarrow$~N}}\\

\midrule
\multirow{2}{*}{\rotatebox[origin=c]{90}{\textbf{\scriptsize{H2O}}}}
&CASA \cite{kwon2022context} &$\blacktriangle$&68.78& 65.15 & 65.09 & 63.69 \\
&\cellcolor{beaublue} LA2DS (Ours) & \cellcolor{beaublue}$\star$ &\cellcolor{beaublue} \textbf{70.12}&\cellcolor{beaublue} \textbf{68.59}  &\cellcolor{beaublue}  \textbf{68.75} &\cellcolor{beaublue} \textbf{68.23} \\
\specialrule{1pt}{1pt}{1pt}
\end{tabular}

}

\caption{Robustness against noisy keypoints. Note that $\blacktriangle$ denotes 3D pose coordinate inputs, while $\star$ denotes 2D pose heatmap inputs. \textbf{C} denotes clean (ground truth) keypoints, while \textbf{N} denotes noisy keypoints.  Best results are in \textbf{bold}.}
%\vspace{-0.5cm}
\label{tab:ablation_keypoint_quality}
\end{minipage}

\end{table}

%% file: Tables/phase_classification.tex
\begin{table}[!t]
\scriptsize
\begin{minipage}[b]{1.0\linewidth}
\centering

{%
\setlength{\tabcolsep}{2pt}
\begin{tabular}{c|c|c|c|c|c}

\specialrule{1pt}{1pt}{1pt}

 & \multirow{2}{*}{\textbf{\scriptsize{Method}}} & \multirow{2}{*}{\textbf{\scriptsize{Input}}} & 
 \multicolumn{3}{c}{\textbf{\scriptsize{Amount of Labels}}} \\ 
 \cline{4-6}
&&&\textbf{\scriptsize{10\%}}&\textbf{\scriptsize{50\%}}&\textbf{\scriptsize{100\%}}\\
\midrule

\multirow{9}{*}{\rotatebox[origin=c]{90}{\textbf{\scriptsize{Penn Action}}}}

& SAL~\cite{ishan2016sal}
&$\square$&79.94&81.11&81.79
\\
& TCN~\cite{sermanet2018tcn}
&$\square$&81.99&82.64&82.78
\\
& TCC~\cite{dwibedi2019tcc} 
&$\square$&79.72&81.12&81.35
 \\
& LAV~\cite{haresh2021learning}
&$\square$&83.56&83.95&84.25
\\
& ViViT~\cite{arnab2021vivit}+LAV~\cite{haresh2021learning}
&$\square$& 70.66 & 72.92 & 74.76 
\\
& VAVA~\cite{liu2022learning}
&$\square$&83.89&84.23&84.48
\\
& CASA~\cite{kwon2022context}
& $\blacktriangle$&\underline{\textit{88.55}}&\underline{\textit{91.87}}&\underline{\textit{92.20}}
\\
& STGCN~\cite{yan2018spatial}+CASA~\cite{kwon2022context}
& $\blacktriangle$& 87.45&90.55&91.37
\\
& \cellcolor{beaublue} LA2DS (Ours)
& \cellcolor{beaublue}$\star$&\cellcolor{beaublue} \textbf{89.27}&\cellcolor{beaublue} \textbf{92.30}&\cellcolor{beaublue} \textbf{92.63}
\\

\midrule
\multirow{7}{*}{\rotatebox[origin=c]{90}{\textbf{\scriptsize{IKEA ASM}}}}

& SAL~\cite{ishan2016sal}
&$\square$&21.68&21.72&22.14
\\
& TCN~\cite{sermanet2018tcn}
&$\square$&25.17&25.70&26.80
\\
& TCC~\cite{dwibedi2019tcc}
&$\square$&24.74&25.22&26.46
\\
& LAV~\cite{haresh2021learning}
&$\square$&\underline{\textit{29.78}}&29.85&30.43
\\
& VAVA~\cite{liu2022learning}
&$\square$&\textbf{31.66}&\textbf{33.79}&\underline{\textit{32.91}}
\\
& CASA~\cite{kwon2022context}
& $\blacktriangle$&21.32&31.52&31.06
\\
& \cellcolor{beaublue} LA2DS (Ours)
& \cellcolor{beaublue}$\star$&\cellcolor{beaublue}26.43&\cellcolor{beaublue} \underline{\textit{32.56}}&\cellcolor{beaublue}\textbf{34.73}
\\

\midrule
\multirow{4}{*}{\rotatebox[origin=c]{90}{\textbf{\scriptsize{H2O}}}}

& TCC~\cite{dwibedi2019tcc}
&$\square$&43.30&52.48&52.78
\\
& LAV~\cite{haresh2021learning}
&$\square$& 35.38 & 51.66 &53.43
\\
& CASA~\cite{kwon2022context}
& $\blacktriangle$&\underline{\textit{43.50}}&\underline{\textit{62.51}}&\underline{\textit{68.78}}
\\
& \cellcolor{beaublue} LA2DS (Ours)
& \cellcolor{beaublue}$\star$&\cellcolor{beaublue} \textbf{52.86}&\cellcolor{beaublue} \textbf{64.58}&\cellcolor{beaublue} \textbf{70.12}

    \\
% - & - & - & 0.6354

\specialrule{1pt}{1pt}{1pt}

\end{tabular}

}

\caption{Phase classification results. Note that $\square$ denotes direct RGB inputs, $\blacktriangle$ denotes 3D pose coordinate inputs, and $\star$ denotes 2D pose heatmap inputs. Best results are in \textbf{bold}, while second best ones are \underline{\textit{underlined}}.}
%\vspace{-0.5cm}
\label{tab:phase_classification}
\end{minipage}

\end{table}

%% file: Tables/progress_tau.tex
\begin{table}[!t]
\scriptsize
\begin{minipage}[b]{1.0\linewidth}
\centering

{%
\setlength{\tabcolsep}{2pt}
\begin{tabular}{c|c|c|c|c}

\specialrule{1pt}{1pt}{1pt}

 & \textbf{\scriptsize{Method}} & \textbf{\scriptsize{Input}} & \textbf{\scriptsize{Progress.}} & \bm{$\tau$}  \\
\midrule

\multirow{9}{*}{\rotatebox[origin=c]{90}{\textbf{\scriptsize{Penn Action}}}}
&SAL \cite{ishan2016sal}&$\square$&0.6960&0.7612 \\
&TCN \cite{sermanet2018tcn}&$\square$&0.7217&0.8120 \\
&TCC \cite{dwibedi2019tcc}&$\square$&0.6638&0.7012 \\
&LAV \cite{haresh2021learning}&$\square$&0.6613&0.8047 \\
&ViViT~\cite{arnab2021vivit}+LAV~\cite{haresh2021learning}&$\square$& 0.7681 & 0.7059 \\
&VAVA \cite{liu2022learning}&$\square$&0.7091&0.8053\\
&CASA \cite{kwon2022context} &$\blacktriangle$&\textbf{0.9449}&\underline{\textit{0.9728}} \\
& STGCN~\cite{yan2018spatial}+CASA~\cite{kwon2022context}& $\blacktriangle$& 0.9061 & 0.9156
\\
&\cellcolor{beaublue} LA2DS (Ours) &\cellcolor{beaublue}$\star$&\cellcolor{beaublue} \underline{\textit{0.9348}}&\cellcolor{beaublue} \textbf{0.9887} \\

\midrule
\multirow{3}{*}{\rotatebox[origin=c]{90}{\textbf{\scriptsize{H2O}}}}
&LAV \cite{haresh2021learning}&$\square$&0.5913&0.5323 \\
&CASA \cite{kwon2022context} &$\blacktriangle$&\underline{\textit{0.9107}}&\underline{\textit{0.9438}} \\
&\cellcolor{beaublue} LA2DS (Ours) &\cellcolor{beaublue}$\star$&\cellcolor{beaublue} \textbf{0.9280}&\cellcolor{beaublue} \textbf{0.9670} \\

\specialrule{1pt}{1pt}{1pt}
\end{tabular}

}

\caption{Phase progression and Kendall’s tau results. Note that $\square$ denotes direct RGB inputs, $\blacktriangle$ denotes 3D pose coordinate inputs, and $\star$ denotes 2D pose heatmap inputs. Best results are in \textbf{bold}, while second best ones are \underline{\textit{underlined}}.}
%\vspace{-0.5cm}
\label{tab:progress_tau}
\end{minipage}

\end{table}

%% file: Tables/frame_retrieval.tex
\begin{table}[!t]
\scriptsize
\begin{minipage}[b]{1.0\linewidth}
\centering

{%
\setlength{\tabcolsep}{2pt}
\begin{tabular}{c|c|c|c|c|c}

\specialrule{1pt}{1pt}{1pt}

 & \textbf{\scriptsize{Method}} & \textbf{\scriptsize{Input}}& \textbf{\scriptsize{AP@5}} & \textbf{\scriptsize{AP@10}} & \textbf{\scriptsize{AP@15}}  \\
\midrule
\multirow{9}{*}{\rotatebox[origin=c]{90}{\textbf{\scriptsize{Penn Action}}}}
&SAL \cite{ishan2016sal}&$\square$&76.04&75.77&75.61\\
&TCN \cite{sermanet2018tcn}&$\square$&77.84&77.51&77.28\\
&TCC \cite{dwibedi2019tcc}&$\square$&76.74&76.27&75.88\\
&LAV \cite{haresh2021learning}&$\square$&79.13&78.98&78.90 \\
&ViViT~\cite{arnab2021vivit}+LAV~\cite{haresh2021learning}&$\square$& 87.58  & 83.07 & 80.16 \\
&VAVA \cite{liu2022learning}&$\square$&81.52&80.47&80.67\\
&CASA \cite{kwon2022context}&$\blacktriangle$&89.90&89.44&\underline{\textit{89.07}}\\
&STGCN~\cite{yan2018spatial}+CASA~\cite{kwon2022context}&$\blacktriangle$&\underline{\textit{90.32}}& \underline{\textit{89.56}} & 88.41\\
&\cellcolor{beaublue} LA2DS (Ours) &\cellcolor{beaublue}$\star$ &\cellcolor{beaublue} \textbf{93.07}&\cellcolor{beaublue} \textbf{91.84}&\cellcolor{beaublue} \textbf{91.35}
\\

\midrule
\multirow{7}{*}{\rotatebox[origin=c]{90}{\textbf{\scriptsize{IKEA ASM}}}}
&SAL \cite{ishan2016sal}&$\square$&15.15&14.90&14.72\\
&TCN \cite{sermanet2018tcn}&$\square$&19.15&19.19&19.33\\
&TCC \cite{dwibedi2019tcc}&$\square$&19.80&19.64&19.68\\
&LAV \cite{haresh2021learning}&$\square$&23.89&23.65&23.56 \\
&VAVA \cite{liu2022learning}&$\square$&\underline{\textit{29.58}}&28.74&28.48\\
&CASA \cite{kwon2022context}&$\blacktriangle$&28.92&\underline{\textit{28.88}}&\underline{\textit{28.61}}\\
&\cellcolor{beaublue}LA2DS (Ours) &\cellcolor{beaublue}$\star$&\cellcolor{beaublue}\textbf{32.44}&\cellcolor{beaublue}\textbf{31.89}&\cellcolor{beaublue}\textbf{31.56}\\

\midrule
\multirow{3}{*}{\rotatebox[origin=c]{90}{\textbf{\scriptsize{H2O}}}}
&LAV \cite{haresh2021learning}&$\square$& 47.55&45.56&44.61\\
&CASA \cite{kwon2022context}&$\blacktriangle$& \underline{\textit{60.13}}& \underline{\textit{59.44}}&\underline{\textit{59.01}}\\
&\cellcolor{beaublue}LA2DS (Ours) &\cellcolor{beaublue}$\star$&\cellcolor{beaublue} \textbf{67.51}&\cellcolor{beaublue} \textbf{63.11}&\cellcolor{beaublue} \textbf{61.75}\\

\specialrule{1pt}{1pt}{1pt}
\end{tabular}

}

\caption{Fine-grained frame retrieval results. Note that $\square$ denotes direct RGB inputs, $\blacktriangle$ denotes 3D pose coordinate inputs, and $\star$ denotes 2D pose heatmap inputs. Best results are in \textbf{bold}, while second best ones are \underline{\textit{underlined}}.}
%\vspace{-0.5cm}
\label{tab:frame_retrieval}
\end{minipage}

\end{table}

%% file: Tables/fusion.tex
\begin{table}[!t]
\scriptsize
\begin{minipage}[b]{1.0\linewidth}
\centering

{%
\setlength{\tabcolsep}{2pt}
\begin{tabular}{c|c|c|c|c|c|c}

\specialrule{1pt}{1pt}{1pt}

 & \textbf{\scriptsize{Method}} & \textbf{\scriptsize{Input}} & \textbf{\scriptsize{Class.}} & \textbf{\scriptsize{Progress.}} & \bm{$\tau$} & \textbf{\scriptsize{Retriev.}}\\
\midrule

\multirow{7}{*}{\rotatebox[origin=c]{90}{\textbf{\scriptsize{Penn Action}}}}
&LAV~\cite{haresh2021learning}
&$\square$  & 84.25 & 0.6613 & 0.8047 & 78.90
\\
&VAVA~\cite{liu2022learning}
&$\square$   & 84.48 & 0.7091 & 0.8053 & 80.67
\\
&CASA~\cite{kwon2022context}
&$\blacktriangle$  & 92.20 & 0.9449 & 0.9728 & 89.07
\\
&CASA~\cite{kwon2022context}
&$\blacktriangle$ + $\square$  & \underline{\textit{92.90}} & \underline{\textit{0.9494}} & 0.9808 & 90.25
\\
&\cellcolor{beaublue}LA2DS (Ours)
&\cellcolor{beaublue}$\star$ & \cellcolor{beaublue}92.63 & \cellcolor{beaublue}0.9348 & \cellcolor{beaublue}\underline{\textit{0.9887}} & \cellcolor{beaublue}\underline{\textit{91.35}}
\\
&\cellcolor{beaublue}LA2DS (Ours)
&\cellcolor{beaublue}$\square$ & \cellcolor{beaublue}88.58 & \cellcolor{beaublue}0.8891 & \cellcolor{beaublue}0.9625 & \cellcolor{beaublue}81.08
\\
&\cellcolor{beaublue}LA2DS (Ours)
&\cellcolor{beaublue}$\star$ + $\square$  & \cellcolor{beaublue}\textbf{93.57} & \cellcolor{beaublue}\textbf{0.9507} & \cellcolor{beaublue}\textbf{0.9903} & \cellcolor{beaublue}\textbf{92.15}
\\

\specialrule{1pt}{1pt}{1pt}
\end{tabular}

}

\caption{Multi-modality fusion results. Note that $\square$ denotes direct RGB inputs, $\blacktriangle$ denotes 3D pose coordinate inputs, and $\star$ denotes 2D pose heatmap inputs. Best results are in \textbf{bold}, while second best ones are \underline{\textit{underlined}}.}
%\vspace{-0.5cm}
\label{tab:fusion}
\end{minipage}

\end{table}

%% file: Sections/conclusion.tex
\section{Conclusion}
\label{sec:conclusion}

We present a 2D skeleton-based self-supervised temporal video alignment framework. Specifically, we propose to take 2D skeleton heatmap inputs and employ video transformers to perform self-attention both in spatial and temporal domains. This is in contrast with CASA, which processes 3D skeleton coordinates and does not conduct spatial self-attention. For self-supervised learning, we exploit simple heatmap augmentation techniques based on 2D skeletons. Our approach achieves both better accuracy and higher robustness than CASA and shows superior results on fine-grained human activity understanding tasks on Penn Action, IKEA ASM, and H2O. Finally, our multi-modality version establishes the state-of-the-art on all metrics and datasets. To our best knowledge, our work is the first to exploit 2D skeleton heatmap inputs and the first to explore multi-modality fusion for temporal video alignment.

%% file: supp.tex
\section{Supplementary Material}

In this supplementary material, we first provide the details of our implementation in Sec.~\ref{sec:implementation}. Next, the ablation results of different augmentations are included in Sec.~\ref{sec:ablation_augmentations}, while our multi-modality model architecture and our complete multi-modality fusion results on all metrics and datasets are presented in Sec.~\ref{sec:fusion_supp}. We then report the ablation results of different method parts in Sec.~\ref{sec:ablation_parts}, while the ablation results of different 2D poses are included in Sec.~\ref{sec:ablation_poses}. Lastly, we show some qualitative results in Sec.~\ref{sec:qualitative} and discuss the limitations and societal impacts of our work in Sec.~\ref{sec:discussions}.

\subsection{Implementation Details}
\label{sec:implementation}

\noindent\textbf{Hyperparameter Settings.}
Tab.~\ref{tab:hyperparams} includes a summary of our hyperparameter settings. We follow the same hyperparameter settings used in CASA~\cite{kwon2022context}, e.g., frames per second, batch size, temperature, learning rate, and weight decay. We keep the augmentation probability the same as CASA~\cite{kwon2022context}. However, we adjust the noise standard deviation to 0.1 radian for orientation augmentation and 1 heatmap pixel for position augmentation, since our augmentation techniques are conducted on 2D skeletons/heatmaps. For CASA~\cite{kwon2022context}, the input vector length increases linearly with the number of keypoints and the number of objects (i.e., humans, hands) involved in each dataset. In contrast, our approach relies on the same heatmap spatial dimension of 30 $\times$ 30 for all datasets, demonstrating better scalability. In addition, for our self-attention module, we follow the same hyperparameter settings used in ViViT~\cite{arnab2021vivit}, except that we reduce the patch size to 10 $\times$ 10 and the transformer output dimension to 192 since our heatmap spatial dimension is small (i.e., 30 $\times$ 30).

\noindent\textbf{Computing Resources.} All of our experiments are performed with a single Nvidia A100 SXM4 GPU on Lambda Cloud. 

\begin{table}[!t]
  \scriptsize
  \centering
  \begin{tabular}{|l|l|}
    \hline
    \textbf{\scriptsize{Hyperparameter}} & \textbf{\scriptsize{Value}} \\ \hline
    Frames per second & 20 (Penn Action), \\ & 30 (H2O, IKEA ASM) \\ \hline
    Batch size & 64 (Penn Action), 32 (H2O), \\ & 4 (IKEA ASM)          \\ \hline
    Temperature & 0.1 \\ \hline
    Learning rate & 3e-3 (Penn Action), 3e-4 (H2O), \\ & 3e-2 (IKEA ASM)    \\ \hline
    Weight decay & 0.5 \\ \hline
    Augmentation probability & 0.3 (flipping, orientation, \\ & position), 0.5 (temporal) \\ \hline
    Noise standard & 0.1 radian (orientation),  \\ deviation & 1 heatmap pixel (position)\\ \hline
    Heatmap spatial & 30 $\times$ 30 \\ dimension & \\ \hline
    Patch size & 10 $\times$ 10 \\ \hline
    Transformer output & 192 \\ dimension &  \\ \hline
    Optimizer & ADAM \\ \hline
    Scheduler & MultiStepLR \\ \hline
    Scheduler interval & Epoch \\ \hline
  \end{tabular}
  \caption{Hyperparameter settings.}
  \label{tab:hyperparams}
\end{table}

\subsection{Effects of Different Augmentations}
\label{sec:ablation_augmentations}

Tab.~\ref{tab:ablation_augmentations} provides the ablation results on various data augmentation techniques. It can be seen from Tab.~\ref{tab:ablation_augmentations} that utilizing all augmentations simultaneously yields the best performance across all metrics, whereas eliminating all augmentations leads to a significant performance drop. While removing orientation augmentation and horizontal flipping causes a drop of roughly 3\% in classification accuracy, removing position augmentation reduces the accuracy by 1\%. In addition, the absence of temporal augmentation has the greatest impact on classification accuracy among all augmentations, underscoring the importance of temporal information.

\input{Tables/ablation_table_augmentations.tex}

\subsection{Multi-Modality Fusion}
\label{sec:fusion_supp}

\begin{figure}[t]
	\centering
		\includegraphics[width=1.0\linewidth, trim = 0mm 0mm 0mm 0mm, clip]{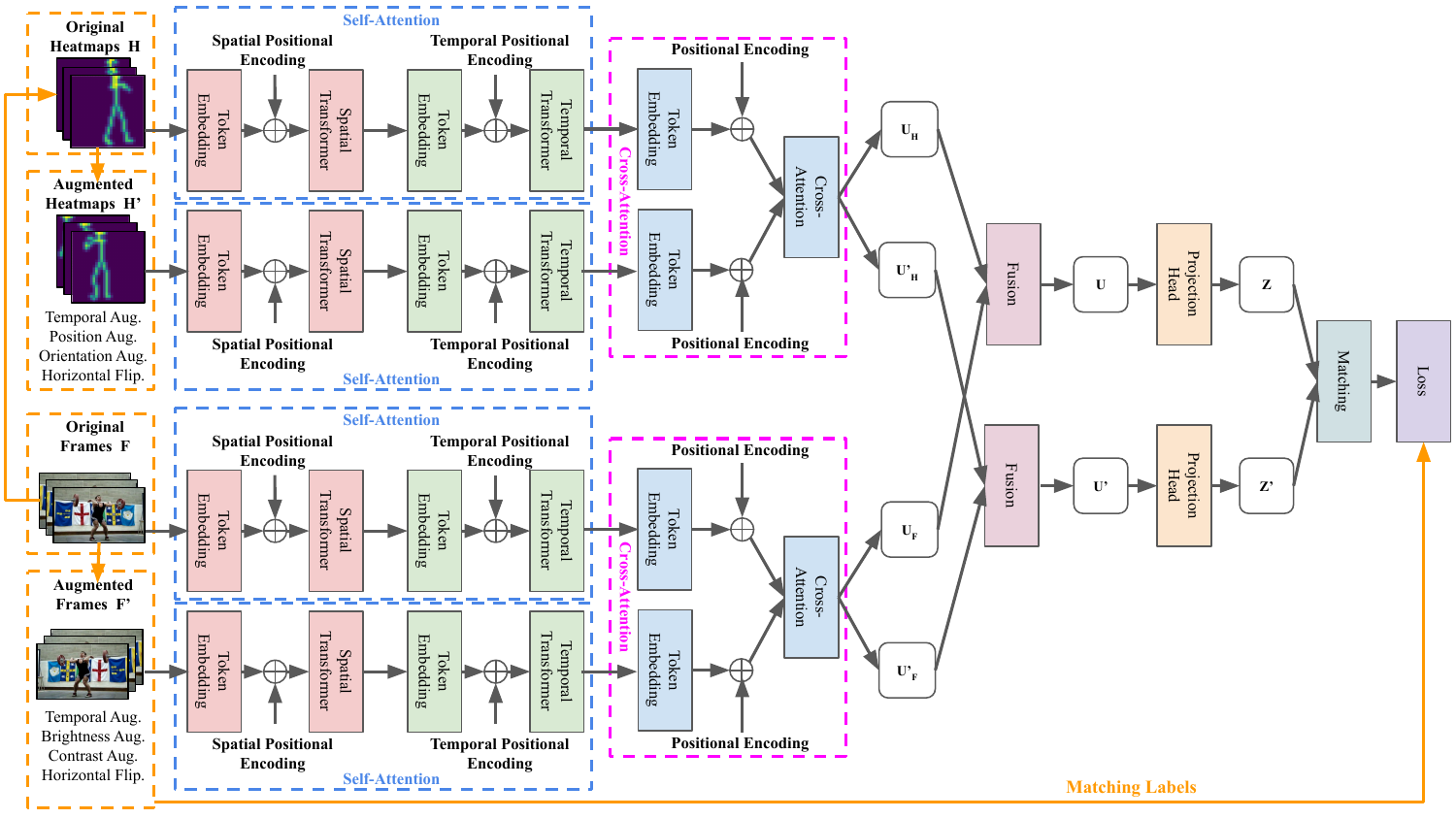}
	\caption{Our multi-modality model includes various encoders to extract features from heatmaps (top) and RGB videos (bottom) and performs late fusion to obtain the combined features.}
	\label{fig:fusion}
\end{figure}

We first illustrate our multi-modality model architecture in Fig.~\ref{fig:fusion}, which we describe in details in Sec.~3.4 of the main paper. Next, our complete multi-modality fusion results on all metrics and datasets are presented in Tab.~\ref{tab:fusion_complete}. It is clear from the results, our RGB+Pose model yields the best numbers across all metrics and datasets, outperforming all the single-modality methods and the multi-modality baseline extended from CASA~\cite{kwon2022context} and establishing the state-of-the-art across all metrics and datasets. In addition, our Pose-only model obtains better results than our RGB-only model on Penn Action and H2O, while having generally worse results on IKEA ASM. It likely implies that pose cues are more important on Penn Action and H2O, while RGB cues are more crucial on IKEA ASM. Finally, both our RGB+Pose model and the CASA-based multi-modality baseline outperform their single-modality counterparts respectively, which validates the benefit of multi-modality inputs for fine-grained human activity understanding tasks.

\begin{table}[!t]
\tiny
\begin{minipage}{\linewidth}
\centering
{%
\setlength{\tabcolsep}{7pt}
\begin{tabular}{c|c|c|c|c|c|c|c|c|c|c}
\specialrule{1pt}{1pt}{1pt}
& \textbf{\tiny{Method}} & \textbf{\tiny{Input}} & \textbf{\tiny{Acc@0.1}} & \textbf{\tiny{Acc@0.5}} & \textbf{\tiny{Acc@1.0}} & \textbf{\tiny{Progress}} & \bm{$\tau$} & \textbf{\tiny{AP@5}} & \textbf{\tiny{AP@10}} & \textbf{\tiny{AP@15}} \\

\midrule
\multirow{7}{*}{\rotatebox[origin=c]{90}{\textbf{\tiny{Penn Action}}}}
&LAV~\cite{haresh2021learning}
&$\square$  & 83.56 & 83.95 & 84.25 & 0.6613 & 0.8047 & 79.13 & 78.98 & 78.90 \\
&VAVA~\cite{liu2022learning}
&$\square$   &  83.89 & 84.23 & 84.48 & 0.7091 & 0.8053 & 81.52 & 80.47 & 80.67\\
&CASA~\cite{kwon2022context}
&$\blacktriangle$  &  88.55 & 91.87 & 92.20 & 0.9449 & 0.9728 & 89.90 & 89.44 & 89.07\\
&CASA~\cite{kwon2022context}
&$\blacktriangle$ + $\square$ & 88.97 & \underline{\textit{92.55}} & \underline{\textit{92.90}} & \underline{\textit{0.9494}} & 0.9808 & 91.74 & 90.97 & 90.25\\
&\cellcolor{beaublue}LA2DS (Ours)
&\cellcolor{beaublue}$\star$ & \cellcolor{beaublue}\underline{\textit{89.27}} & \cellcolor{beaublue}92.30 & \cellcolor{beaublue}92.63 & \cellcolor{beaublue}0.9348 & \cellcolor{beaublue}\underline{\textit{0.9887}} & \cellcolor{beaublue}\underline{\textit{93.07}} & \cellcolor{beaublue}\underline{\textit{91.84}} & \cellcolor{beaublue}\underline{\textit{91.35}} \\
&\cellcolor{beaublue}LA2DS (Ours)
&\cellcolor{beaublue}$\square$ & \cellcolor{beaublue}84.75 & \cellcolor{beaublue}86.91 & \cellcolor{beaublue}88.58 & \cellcolor{beaublue}0.8891 & \cellcolor{beaublue}0.9625 & \cellcolor{beaublue}84.81 & \cellcolor{beaublue}82.16 & \cellcolor{beaublue}81.08\\
&\cellcolor{beaublue}LA2DS (Ours)
&\cellcolor{beaublue}$\star$ + $\square$  & \cellcolor{beaublue}\textbf{90.22} & \cellcolor{beaublue}\textbf{93.02} & \cellcolor{beaublue}\textbf{93.57} & \cellcolor{beaublue}\textbf{0.9507} & \cellcolor{beaublue}\textbf{0.9903} & \cellcolor{beaublue}\textbf{93.63} & \cellcolor{beaublue}\textbf{92.25} & \cellcolor{beaublue}\textbf{92.15} \\

\midrule
\multirow{7}{*}{\rotatebox[origin=c]{90}{\textbf{\tiny{IKEA ASM}}}}
&LAV~\cite{haresh2021learning}
&$\square$  &  29.78 & 29.85 & 30.43 & -  & - & 23.89  & 23.65 & 23.56\\
&VAVA~\cite{liu2022learning}
&$\square$   &   \underline{\textit{31.66}} & \underline{\textit{33.79}} & 32.91 & - & - & 29.58  & 28.74 & 28.48\\
&CASA~\cite{kwon2022context}
&$\blacktriangle$  &   21.32 & 31.52 & 31.06 & - & -  & 28.92  & 28.88  & 28.61\\
&CASA~\cite{kwon2022context}
&$\blacktriangle$ + $\square$  & 27.63 & 32.85 & 34.33 &  &  & 31.45 & 31.08 & 30.87 \\
&\cellcolor{beaublue}LA2DS (Ours)
&\cellcolor{beaublue}$\star$ & \cellcolor{beaublue}26.43 & \cellcolor{beaublue}32.56 & \cellcolor{beaublue}34.73 & \cellcolor{beaublue}- & \cellcolor{beaublue}- & \cellcolor{beaublue}32.44 & \cellcolor{beaublue}\underline{\textit{31.89}} & \cellcolor{beaublue}31.56\\
&\cellcolor{beaublue}LA2DS (Ours)
&\cellcolor{beaublue}$\square$ & \cellcolor{beaublue}28.31 & \cellcolor{beaublue}32.98 & \cellcolor{beaublue}\underline{\textit{35.55}} & \cellcolor{beaublue}- & \cellcolor{beaublue}- & \cellcolor{beaublue}\underline{\textit{32.65}} & \cellcolor{beaublue}31.63 & \cellcolor{beaublue}\underline{\textit{31.60}}\\
&\cellcolor{beaublue}LA2DS (Ours)
&\cellcolor{beaublue}$\star$ + $\square$  & \cellcolor{beaublue}\textbf{31.74} & \cellcolor{beaublue}\textbf{34.51} & \cellcolor{beaublue}\textbf{36.18} & \cellcolor{beaublue}- & \cellcolor{beaublue}- & \cellcolor{beaublue}\textbf{33.08} & \cellcolor{beaublue}\textbf{32.25} & \cellcolor{beaublue}\textbf{32.17} \\

\midrule
\multirow{5}{*}{\rotatebox[origin=c]{90}{\textbf{\tiny{H2O}}}}
&LAV~\cite{haresh2021learning}
&$\square$  &   35.38 & 51.66 & 53.43 & 0.5913 & 0.5323 & 47.55 & 45.56 & 44.61\\
&CASA~\cite{kwon2022context}
&$\blacktriangle$  & 43.50 & 62.51 & 68.78 & 0.9107 & 0.9438 & 60.13 & 59.44 & 59.01\\
&\cellcolor{beaublue}LA2DS (Ours)
&\cellcolor{beaublue}$\star$ & \cellcolor{beaublue}\underline{\textit{52.86}} & \cellcolor{beaublue}\underline{\textit{64.58}} & \cellcolor{beaublue}\underline{\textit{70.12}} & \cellcolor{beaublue}\underline{\textit{0.9280}} & \cellcolor{beaublue}\underline{\textit{0.9670}} & \cellcolor{beaublue}\underline{\textit{67.51}} & \cellcolor{beaublue}\underline{\textit{63.11}} & \cellcolor{beaublue}\underline{\textit{61.75}}\\
&\cellcolor{beaublue}LA2DS (Ours)
&\cellcolor{beaublue}$\square$ & \cellcolor{beaublue}48.08 & \cellcolor{beaublue}63.26 & \cellcolor{beaublue}68.85 & \cellcolor{beaublue}0.9113 & \cellcolor{beaublue}0.9478 & \cellcolor{beaublue}63.33 & \cellcolor{beaublue}61.87 & \cellcolor{beaublue}60.12\\
&\cellcolor{beaublue}LA2DS (Ours)
&\cellcolor{beaublue}$\star$ + $\square$  & \cellcolor{beaublue}\textbf{54.55} & \cellcolor{beaublue}\textbf{65.96} & \cellcolor{beaublue}\textbf{70.65} & \cellcolor{beaublue}\textbf{0.9367} & \cellcolor{beaublue}\textbf{0.9737} & \cellcolor{beaublue}\textbf{70.33} & \cellcolor{beaublue}\textbf{67.76} & \cellcolor{beaublue}\textbf{63.23} \\

\specialrule{1pt}{1pt}{1pt}
\end{tabular}
}
\caption{Multi-modality fusion results. Note that $\square$ denotes direct RGB inputs, $\blacktriangle$ denotes 3D pose coordinate inputs, and $\star$ denotes 2D pose heatmap inputs. Best results are in \textbf{bold}, while second best ones are \underline{\textit{underlined}}.}
\label{tab:fusion_complete}
\end{minipage}
\end{table}

\subsection{Effects of Different Method Parts}
\label{sec:ablation_parts}

To analyze sources of performance gains, we conduct an ablation study where various parts of our method, i.e., 2D skeleton heatmap inputs + 2D skeleton heatmap augmentations, spatial transfer, and multi-modality fusion, are enabled. Tab.~\ref{tab:ablation_parts} presents the results of CASA~\cite{kwon2022context} (top row) and our variants. In particular, the second row shows the results of our variant ($\heartsuit$), where we disable our 2D skeleton heatmap inputs+2D skeleton heatmap augmentations (and use the 3D skeleton coordinate inputs+3D skeleton coordinate augmentations of CASA~\cite{kwon2022context} instead) and our multi-modality fusion, and further replace the spatial image transformer of ViViT~\cite{arnab2021vivit} with the spatial graph transformer of ST-TR~\cite{plizzari2021skeleton} due to 3D skeleton coordinate inputs. Next, the third row shows the results of our variant ($\diamondsuit$), where we enable our 2D skeleton heatmap inputs+2D skeleton heatmap augmentations, but disable our spatial transformer (and use MLPs instead) and our multi-modality fusion. Lastly, the forth row shows the results of our variant without multi-modality fusion ($\spadesuit$), while the bottom row shows the results of our complete method ($\clubsuit$). 

From the results in Tab.~\ref{tab:ablation_parts}, using either 2D skeleton heatmap inputs+2D skeleton heatmap augmentations ($\heartsuit$) or spatial transformer ($\diamondsuit$) yields minor performance gains over CASA~\cite{kwon2022context}. Moreover, using both 2D skeleton heatmap inputs+2D skeleton heatmap augmentations and spatial transformer ($\spadesuit$) leads to major performance gains over CASA~\cite{kwon2022context}. Finally, adding multi-modality fusion ($\clubsuit$) boosts our results further.

\input{Tables/ablation_table_parts.tex}

\subsection{Effects of Different 2D Poses}
\label{sec:ablation_poses}

\begin{figure}[!t]
	\centering
		\includegraphics[width=0.5\linewidth, trim = 0mm 55mm 110mm 0mm, clip]{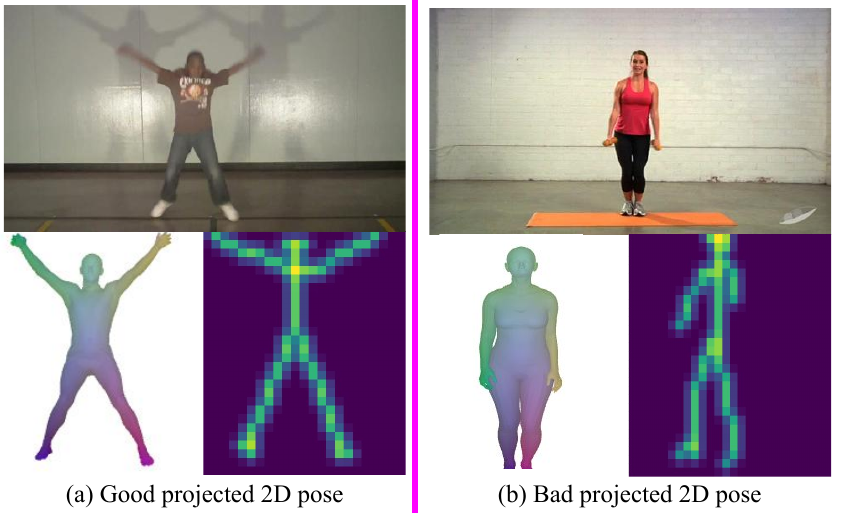}
	\caption{Examples of our projected 2D poses. In (a) or (b), RGB frame is plotted at the top, while 3D pose provided by CASA and our projected 2D pose are visualized at the bottom.}
	\label{fig:projected_poses}
\end{figure}

In all of the previous experiments, we have used 2D poses that are projected (via orthographic projection) from 3D poses provided by CASA~\cite{kwon2022context} for fair comparison purposes. We use orthographic projection since camera intrinsics/extrinsics are not available for all datasets (e.g., Penn Action). In this section, we study the impacts of different 2D poses on the performance of our approach. Specifically, we use estimated 2D poses (computed by OpenPose~\cite{cao2019openpose}) on Penn Action and IKEA ASM and ground truth 2D poses on IKEA ASM (ground truth 2D poses are not available on Penn Action). The ablation results of different 2D poses are shown in Tab.~\ref{tab:ablation_poses}. It can be seen that that using OpenPose 2D poses generally yields better performance than using our projected 2D poses and further advances the state-of-the-art in this topic. This is likely because our projected 2D poses have deformations/deviations due to orthographic projection. Examples of our projected 2D poses are plotted in Fig.~\ref{fig:projected_poses}. Furthermore, using ground truth 2D poses yields the best results, outperforming using OpenPose 2D poses and using our projected 2D poses.

\begin{table}[!t]
\tiny
\begin{minipage}{\linewidth}
\centering
{%
\setlength{\tabcolsep}{7pt}
\begin{tabular}{c|c|c|c|c|c|c|c|c|c|c}
\specialrule{1pt}{1pt}{1pt}
& \textbf{\tiny{Method}} & \textbf{\tiny{Input}} & \textbf{\tiny{Acc@0.1}} & \textbf{\tiny{Acc@0.5}} & \textbf{\tiny{Acc@1.0}} & \textbf{\tiny{Progress}} & \bm{$\tau$} & \textbf{\tiny{AP@5}} & \textbf{\tiny{AP@10}} & \textbf{\tiny{AP@15}} \\

\midrule
\multirow{4}{*}{\rotatebox[origin=c]{90}{\textbf{\tiny{Penn Action}}}}
&Projected
&$\star$ & 89.27 & 92.30 & 92.63 & 0.9348 & 0.9887 & 93.07 & 91.84 & 91.35 \\
&Projected
&$\star$ + $\square$  & 90.22 & 93.02 & 93.57 & 0.9507 & 0.9903 & 93.63 & 92.25 & 92.15 \\
&OpenPose
&$\star$ & \underline{\textit{96.41}} & \underline{\textit{96.81}} & \underline{\textit{96.95}} & \underline{\textit{0.9563}} & \underline{\textit{0.9905}} & \underline{\textit{95.65}} & \underline{\textit{95.43}} & \underline{\textit{95.34}} \\
&OpenPose
&$\star$ + $\square$ & \textbf{96.87} & \textbf{97.22} & \textbf{97.54} & \textbf{0.9633} & \textbf{0.9906} & \textbf{95.98} & \textbf{96.27} & \textbf{96.44} \\

\midrule
\multirow{6}{*}{\rotatebox[origin=c]{90}{\textbf{\tiny{IKEA ASM}}}}
&Projected
&$\star$ & 26.43 & 32.56 & 34.73 & - & - & 32.44 & 31.89 & 31.56\\
&Projected
&$\star$ + $\square$  & 31.74 & 34.51 & 36.18 & - & - & 33.08 & 32.25 & 32.17 \\
&OpenPose
&$\star$ & 26.78 & 32.79 & 34.41 & - & - & 33.49 & 33.04 & 32.87 \\
&OpenPose
&$\star$ + $\square$  & \underline{\textit{32.13}} & \underline{\textit{34.76}} & \underline{\textit{36.33}} & - & - & 33.51 & 33.07 & 32.97 \\
&Ground Truth
&$\star$ & 27.32 & 34.39 & 35.86 & - & - & \underline{\textit{34.61}} & \underline{\textit{33.69}} & \underline{\textit{33.49}} \\
&Ground Truth
&$\star$ + $\square$ & \textbf{34.52} & \textbf{36.71} & \textbf{38.83} &  &  & \textbf{35.49} & \textbf{34.04} & \textbf{33.82} \\

\specialrule{1pt}{1pt}{1pt}
\end{tabular}
}
\caption{Effects of different 2D poses. Note that $\square$ denotes direct RGB inputs, $\blacktriangle$ denotes 3D pose coordinate inputs, and $\star$ denotes 2D pose heatmap inputs. Best results are in \textbf{bold}, while second best ones are \underline{\textit{underlined}}.}
\label{tab:ablation_poses}
\end{minipage}
\end{table}

\subsection{Qualitative Results}
\label{sec:qualitative}

\begin{figure}[!t]
	\centering
	\begin{subfigure}[b]{\linewidth}
		\centering
		\includegraphics[width=1.0\linewidth, clip, page=1]{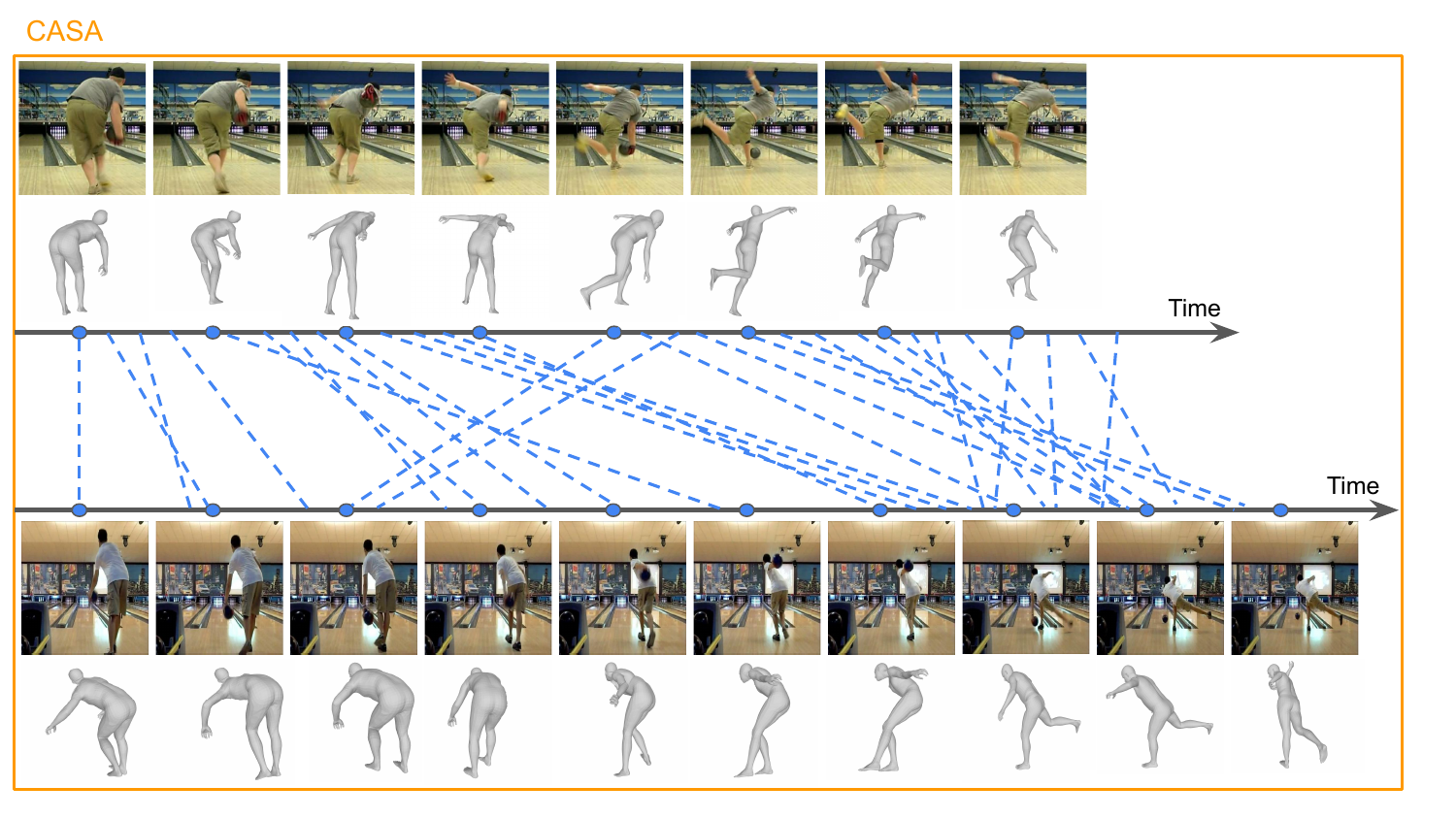}
	\end{subfigure}
	\begin{subfigure}[b]{\linewidth}
		\centering
		\includegraphics[width=1.0\linewidth, clip, page=2]{Figures/alignment_results_2.pdf}
	\end{subfigure}
	\caption{Alignment results of two \textit{bowling} sequences. Blue lines indicate where frames in both sequences match.}
	\label{fig:alignment2}
\end{figure}

\begin{figure}[!t]
	\centering
	\begin{subfigure}[b]{\linewidth}
		\centering
		\includegraphics[width=1.0\linewidth, clip, page=1]{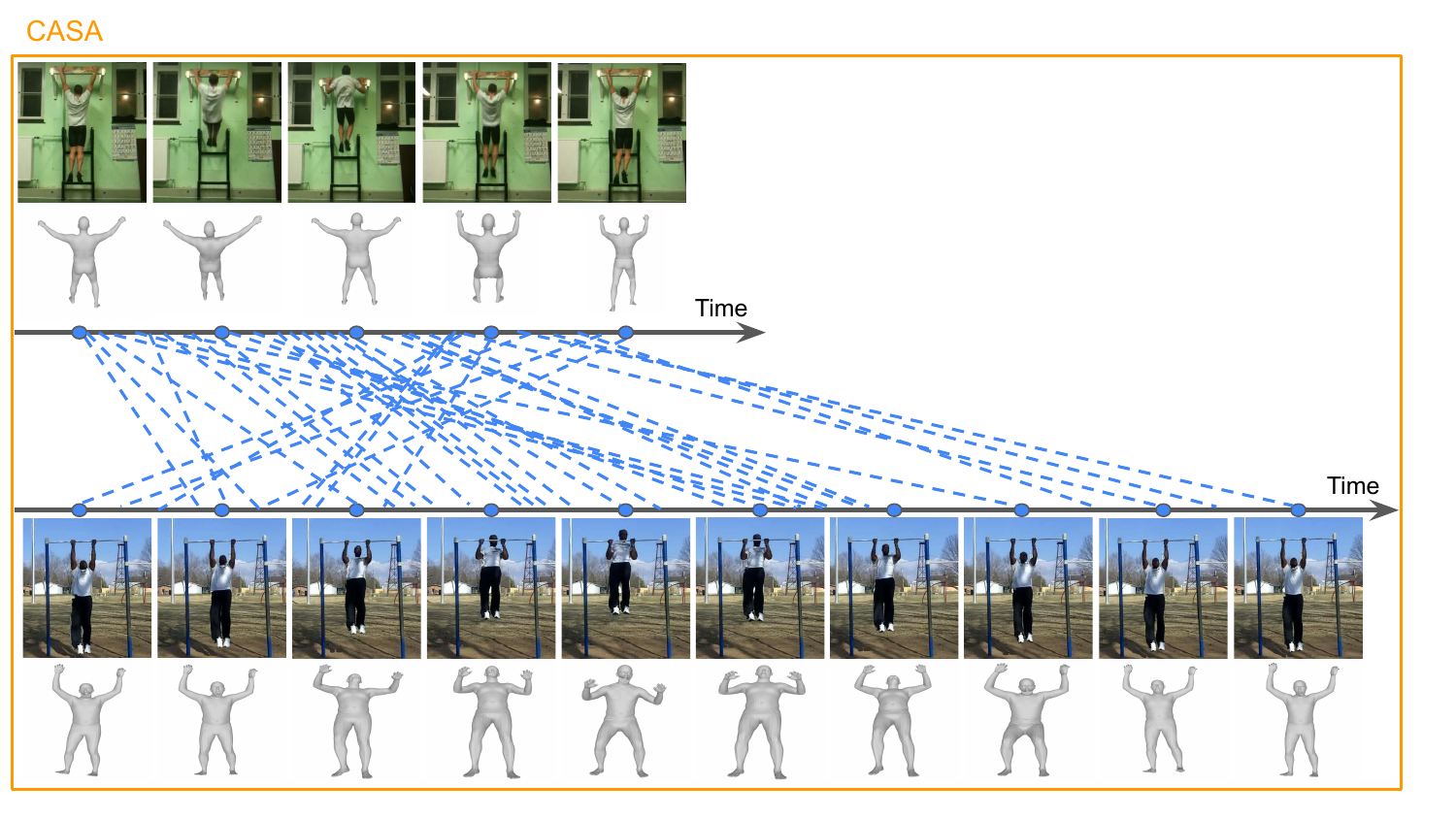}
	\end{subfigure}
	\begin{subfigure}[b]{\linewidth}
		\centering
		\includegraphics[width=1.0\linewidth, clip, page=2]{Figures/alignment_results_1.pdf}
	\end{subfigure}
	\caption{Alignment results of two \textit{pullups} sequences. Blue lines indicate where frames in both sequences match.}
	\label{fig:alignment1}
\end{figure}

Firstly, Figs.~\ref{fig:alignment2} and~\ref{fig:alignment1} illustrate the alignment results of our approach and CASA~\cite{kwon2022context} between two \textit{bowling} sequences and two \textit{pullups} sequences. It is evident from the figures that LA2DS achieves more seamless alignment results than CASA~\cite{kwon2022context}. Our method learns interactions between joints in the same skeleton and between skeletons in the same sequence and across the sequences to align them in time effectively. 

\begin{figure}[!t]
	\centering
	\begin{subfigure}[b]{\linewidth}
		\centering
		\includegraphics[width=1.0\linewidth, clip, page=1]{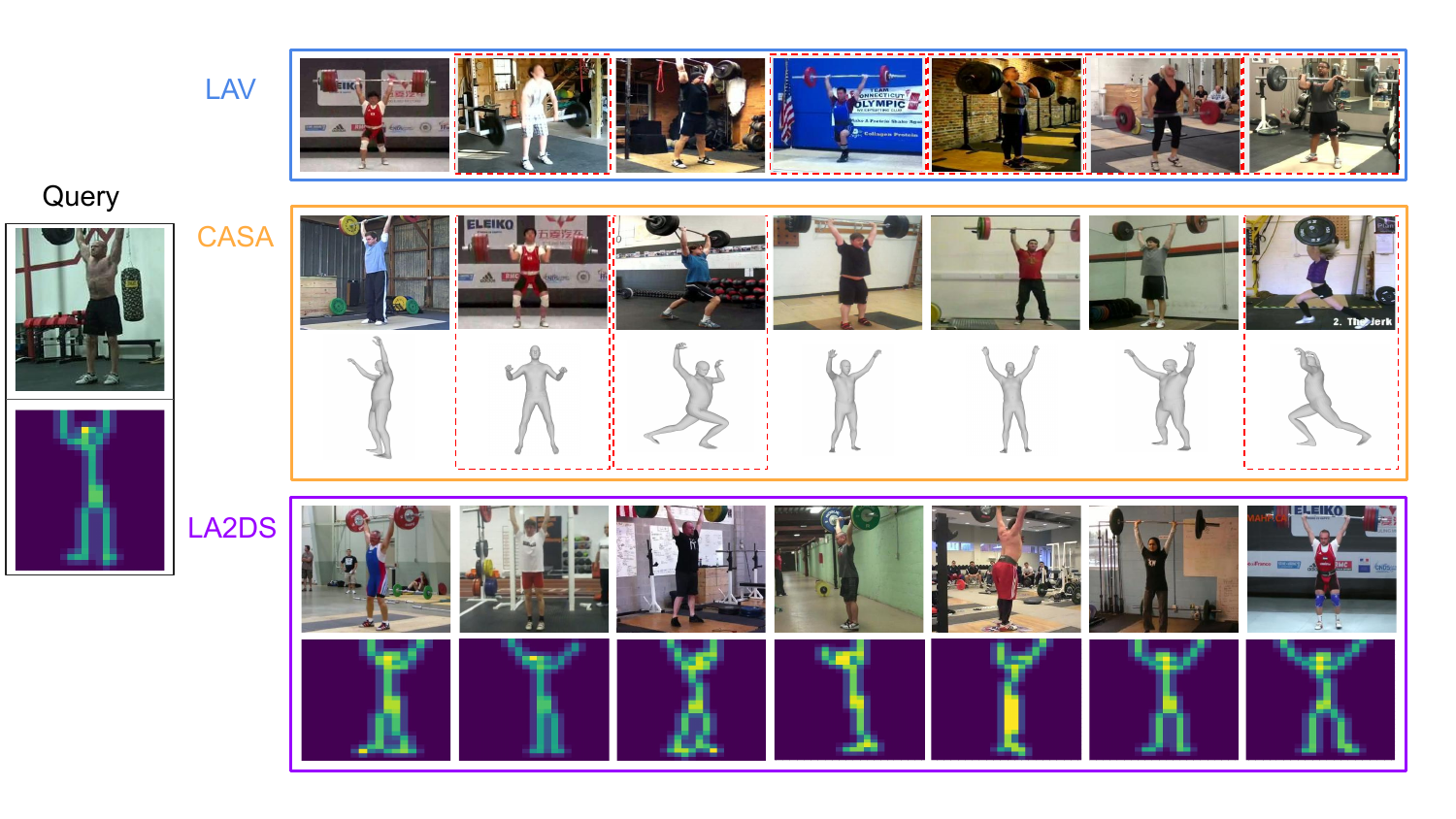}
		\label{fig:retrieval1}
	\end{subfigure}
	\begin{subfigure}[b]{\linewidth}
		\centering
		\includegraphics[width=1.0\linewidth,  clip, page=2]{Figures/retrieval_result.pdf}
		\label{fig:retrieval2}
	\end{subfigure}
	\caption{Fine-grained frame retrieval results on \textit{clean\_and\_jerk} and \textit{jumping\_jacks} sequences with $k=7$. Mismatches between the action phase of the query frame and those of the retrieved frames are indicated by red boxes.}
	\label{fig:retrieval}
\end{figure}

Moreover, Fig.~\ref{fig:retrieval} shows the fine-grained frame retrieval results of our approach, CASA~\cite{kwon2022context}, and LAV~\cite{haresh2021learning} on \textit{clean\_and\_jerk} and \textit{jumping\_jacks} sequences. LA2DS extracts useful spatiotemporal and contextual features within each sequence, resulting in more accurate fine-grained frame retrieval results than CASA~\cite{kwon2022context} and LAV~\cite{haresh2021learning}, as can be seen in Fig.~\ref{fig:retrieval}.

\begin{figure}[!t]
	\centering
		\includegraphics[width=0.7\linewidth, trim = 0mm 50mm 55mm 0mm, clip]{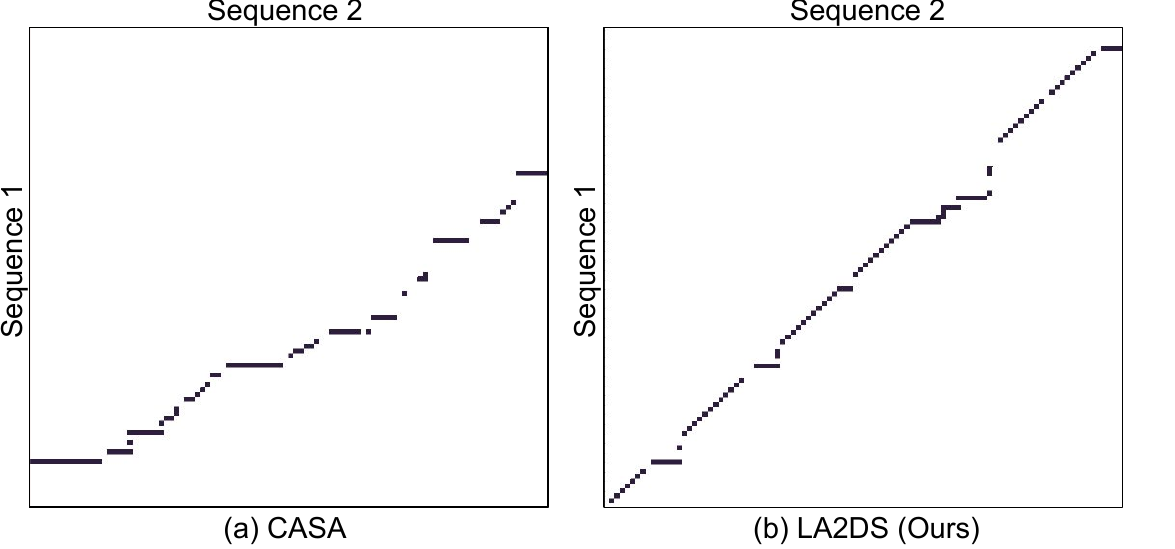}
	\caption{Alignment results of two \emph{tennis\_serve} videos. Black dots indicate alignment results.}
	\label{fig:alignment}
\end{figure}

\begin{figure}[!t]
	\centering
		\includegraphics[width=0.7\linewidth, trim = 0mm 55mm 65mm 0mm, clip]{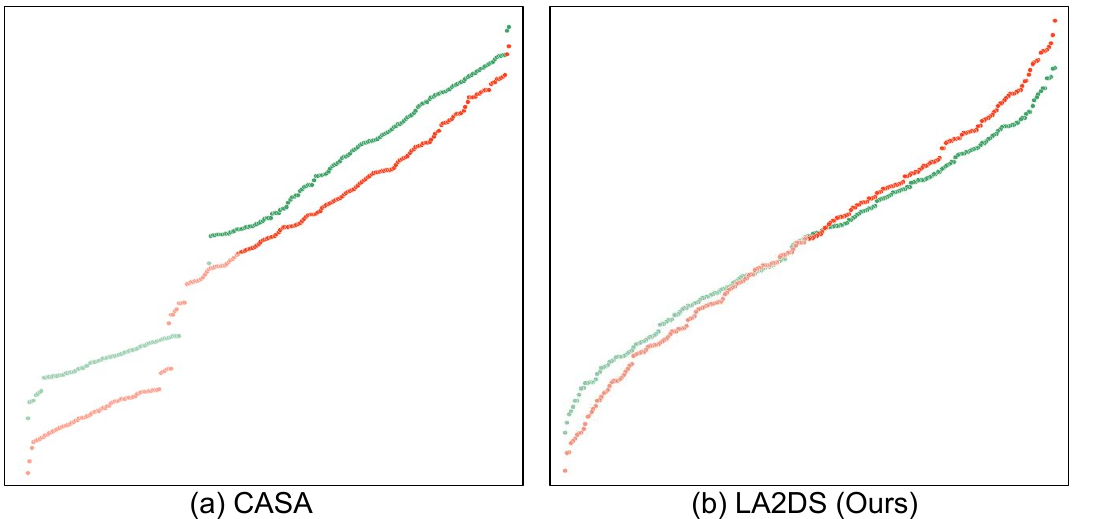}
	\caption{t-SNE visualizations of learned embeddings of two \emph{baseball\_pitch} videos. The opacity of the color denotes the temporal frame index from the first to the last.}
	\label{fig:tsne}
\end{figure}

Finally, we show the alignment results, i.e., framewise correspondences, between two \textit{tennis\_serve} sequences in Fig.~\ref{fig:alignment}, and the t-SNE~\cite{tsne2008} visualization of the learned embeddings of two \textit{baseball\_pitch} sequences in Fig.~\ref{fig:tsne}. Our method achieves a more seamless alignment as compared to CASA~\cite{kwon2022context}. 


\subsection{Discussions}
\label{sec:discussions}

\subsection{Limitations}

\noindent\textbf{Challenging Viewpoints and Extreme Viewpoint Gaps.} 
We follow all of previous works to use Penn Action, IKEA ASM, and H2O, which contain frontal, side, or top viewpoints of human actions (e.g., Fig.~5 of the main paper), since it is difficult to capture human actions with rear or bottom viewpoints due to severe occlusions or privacy concerns respectively. Also, it is generally harder to obtain accurate 3D poses than accurate 2D poses, especially for challenging viewpoints, thus CASA~\cite{kwon2022context}'s sensitivity to noise/errors of pose inputs may further be amplified with challenging viewpoints. However, we do randomly pair videos to perform alignment, yielding viewpoint gaps of up to $180$ degrees (e.g., pairing videos with left and right side viewpoints in Figs.~5(a-b) of the main paper). Collecting a human action dataset including rear or bottom viewpoints and learning by aligning videos with extreme viewpoint gaps are challenging and remain our future works.

\noindent\textbf{Missing Context Details.}
Both 3D skeleton inputs (e.g., CASA~\cite{kwon2022context}) and 2D skeleton inputs (i.e., our method) do not include context details, e.g., objects and background information, which are available in RGB video inputs. Thus, our method may fail in cases where context details are important in recognizing the actions. Nevertheless, our multi-modality fusion method can overcome this limitation by utilizing both 2D skeleton heatmaps and RGB videos as inputs.

\subsection{Societal Impacts}
Our sequence alignment method offers valuable applications including frontline worker training and assistance. In particular, frontline workers could benefit from guidance provided by models generated from expert demonstration videos in various domains from factory assembly to medical surgery. However, we also acknowledge the potential for misuse in surveillance and monitoring of individuals. This may raise privacy concerns, emphasizing the importance of responsible AI principles to guide the use of this technology.

%% file: Tables/ablation_table_augmentations.tex
\begin{table}[!t]
\scriptsize
\begin{minipage}[b]{1.0\linewidth}
\centering

{%
\setlength{\tabcolsep}{2pt}
\begin{tabular}{c|c|c|c|c}

\specialrule{1pt}{1pt}{1pt}

 & \textbf{\scriptsize{Method}} & \textbf{\scriptsize{Class.}} & \textbf{\scriptsize{Progress.}} & \bm{$\tau$}\\
\midrule

\multirow{6}{*}{\rotatebox[origin=c]{90}{\textbf{\scriptsize{H2O}}}}
&No Aug.
& 63.12&0.9080&0.9350

  \\
 &w/o Temp. Aug.
& 64.82&0.9194&0.9551

\\
&w/o Ori. Aug.
& 67.40&0.9203&0.9603

\\
&w/o Pos. Aug.
&\underline{\textit{69.18}}&\underline{\textit{0.9228}}&0.9626

\\
&w/o Hor. Flip.
& 67.93&0.9193&\underline{\textit{0.9642}}

\\
&All
&\textbf{70.12}    &\textbf{0.9280}  &\textbf{0.9670}

\\

\specialrule{1pt}{1pt}{1pt}
\end{tabular}

}

\caption{Effects of different augmentations. Best results are in \textbf{bold}, while second best ones are \underline{\textit{underlined}}.}
\label{tab:ablation_augmentations}
\end{minipage}

\end{table}

%% file: Tables/ablation_table_parts.tex
\begin{table}[!t]
\scriptsize
\begin{minipage}[b]{1.0\linewidth}
\centering
\scriptsize
{%
\setlength{\tabcolsep}{2pt}
\begin{tabular}{c|c|c|c|c|c|c|c}

\specialrule{1pt}{1pt}{1pt}

&\textbf{\scriptsize{Method}} &\textbf{\scriptsize{H+A}} &\textbf{\scriptsize{S}} &\textbf{\scriptsize{F}} &\textbf{\scriptsize{Class.}} &\textbf{\scriptsize{Progress.}} &\bm{$\tau$}\\
\midrule

\multirow{5}{*}{\rotatebox[origin=c]{90}{\textbf{\scriptsize{H2O}}}}
&CASA~\cite{kwon2022context}  &$\times$ &$\times$ &$\times$ &68.78 &0.9107 &0.9438 \\
&LA2DS$\heartsuit$ &$\times$ &\checkmark &$\times$ &68.95 &0.9152 &0.9493 \\
&LA2DS$\diamondsuit$ &\checkmark &$\times$ &$\times$ &68.47 &0.9039 &0.9548 \\
&LA2DS$\spadesuit$ &\checkmark &\checkmark &$\times$ &\underline{\textit{70.12}} &\underline{\textit{0.9280}} &\underline{\textit{0.9670}} \\
&LA2DS$\clubsuit$ &\checkmark &\checkmark &\checkmark &\textbf{70.65} &\textbf{0.9367} &\textbf{0.9737} \\

\specialrule{1pt}{1pt}{1pt}
\end{tabular}
}

\caption{Effects of different method parts, including 2D skeleton heatmap inputs+2D skeleton heatmap augmentations (\textbf{H+A}), spatial transformer (\textbf{S}), and multi-modality fusion (\textbf{F}).}
\label{tab:ablation_parts}
\end{minipage}

\end{table}